\documentclass[lettersize,journal]{IEEEtran}
\usepackage{amsmath,amsfonts}
\usepackage{algorithmic}
\usepackage{algorithm}
\usepackage{array}
\usepackage[caption=false,font=normalsize,labelfont=sf,textfont=sf]{subfig}
\usepackage{textcomp}
\usepackage{stfloats}
\usepackage{url}
\usepackage{verbatim}
\usepackage{graphicx}
\usepackage{booktabs}
\usepackage{colortbl}
\usepackage{multirow}
\usepackage{float}
\usepackage[pagebackref,breaklinks,colorlinks,citecolor=black]{hyperref}

\def\BibTeX{{\rm B\kern-.05em{\sc i\kern-.025em b}\kern-.08em
    T\kern-.1667em\lower.7ex\hbox{E}\kern-.125emX}}
\usepackage{balance}

\def\eg{\emph{e.g.}}

\def\ie{\emph{i.e.}}

\definecolor{hl}{RGB}{240,240,240}

\definecolor{highlightblue}{RGB}{10,60,230}

\usepackage{amsmath} 
\usepackage{bm}      
\begin{document}
\title{Forensics Adapter: Unleashing CLIP for Generalizable Face Forgery Detection}
\author{
\thanks{This work was supported by the National Natural Science Foundation
of China (No.62402464) and the Shandong Natural Science
Foundation (No.ZR2024QF035). The work of Jiaran Zhou was supported by the Shandong Natural Science Foundation (No.ZR2024MF083). \textit{(Corresponding author: Yuezun Li.)}}
Xinjie Cui, Yuezun Li,~\IEEEmembership{Senior Member,~IEEE}, Delong Zhu, Jiaran Zhou,~\IEEEmembership{Member,~IEEE}, Junyu Dong,~\IEEEmembership{Member,~IEEE}, Siwei Lyu,~\IEEEmembership{Fellow,~IEEE} \thanks{Xinjie Cui, Yuezun Li, Delong Zhu, Jiaran Zhou, and Junyu Dong are with the school of Computer Science and Technology, Ocean University of China, Qingdao 266100, China (emails: cuixinjie@stu.ouc.edu.cn; liyuezun@ouc.edu.cn; zhudelong@stu.ouc.edu.cn; zhoujiaran@ouc.edu.cn; dongjunyu@ouc.edu.cn).}
\thanks{Siwei Lyu is with the University at Buffalo, SUNY, USA. {(email: siweilyu@buffalo.edu)}.}
}

\markboth{Journal of \LaTeX\ Class Files,~Vol.~18, No.~9, September~2020}%
{How to Use the IEEEtran \LaTeX \ Templates}
\maketitle
\begin{abstract}
We describe Forensics Adapter, an adapter network designed to transform CLIP into an effective and generalizable face forgery detector. Although CLIP is highly versatile, adapting it for face forgery detection is non-trivial as forgery-related knowledge is entangled with a wide range of unrelated knowledge. Existing methods treat CLIP merely as a feature extractor, lacking task-specific adaptation, which limits their effectiveness. To address this, we introduce an adapter to learn face forgery traces -- the blending boundaries unique to forged faces, guided by task-specific objectives. Then we enhance the CLIP visual tokens with a dedicated interaction strategy that communicates knowledge across CLIP and the adapter. Since the adapter is alongside CLIP, its versatility is highly retained, naturally ensuring strong generalizability in face forgery detection. {With only $\bm{5.7M}$ trainable parameters, our method achieves superior performance across six standard datasets.} Additionally, we describe Forensics Adapter++, an extended method that incorporates textual modality via a newly proposed forgery-aware prompt learning strategy. This extension leads to a further $\bm{1.3\%}$ performance boost over the original Forensics Adapter. We believe the proposed methods can serve as a baseline for future CLIP-based face forgery detection methods. The code has been released at \url{https://github.com/OUC-VAS/ForensicsAdapter}.

\end{abstract}

\section{Introduction}
\label{sec:intro}

\IEEEPARstart{F}{ace} forgery techniques\footnote{Face forgery strictly refers to techniques that manipulate local facial content, such as altering lip movement, swapping central face, editing expression, etc. Whole facial image synthesis using GAN or Diffusion models is therefore not considered in this work.} have seen remarkable progress in recent years, largely due to significant advancements in generative models~\cite{gan_ori,gan_style,difussion_ori,diffusion_stable}. 
{These techniques enable the manipulation of faces with high realism, raising serious societal concerns, such as privacy violations~\cite{privacy1,Privacy2} and economic fraud~\cite{economic1,economic2}.}
These concerns have motivated a great need for exploration into face forgery detection.

\begin{figure}[t]
    \centering
    \includegraphics[width=0.8\linewidth]{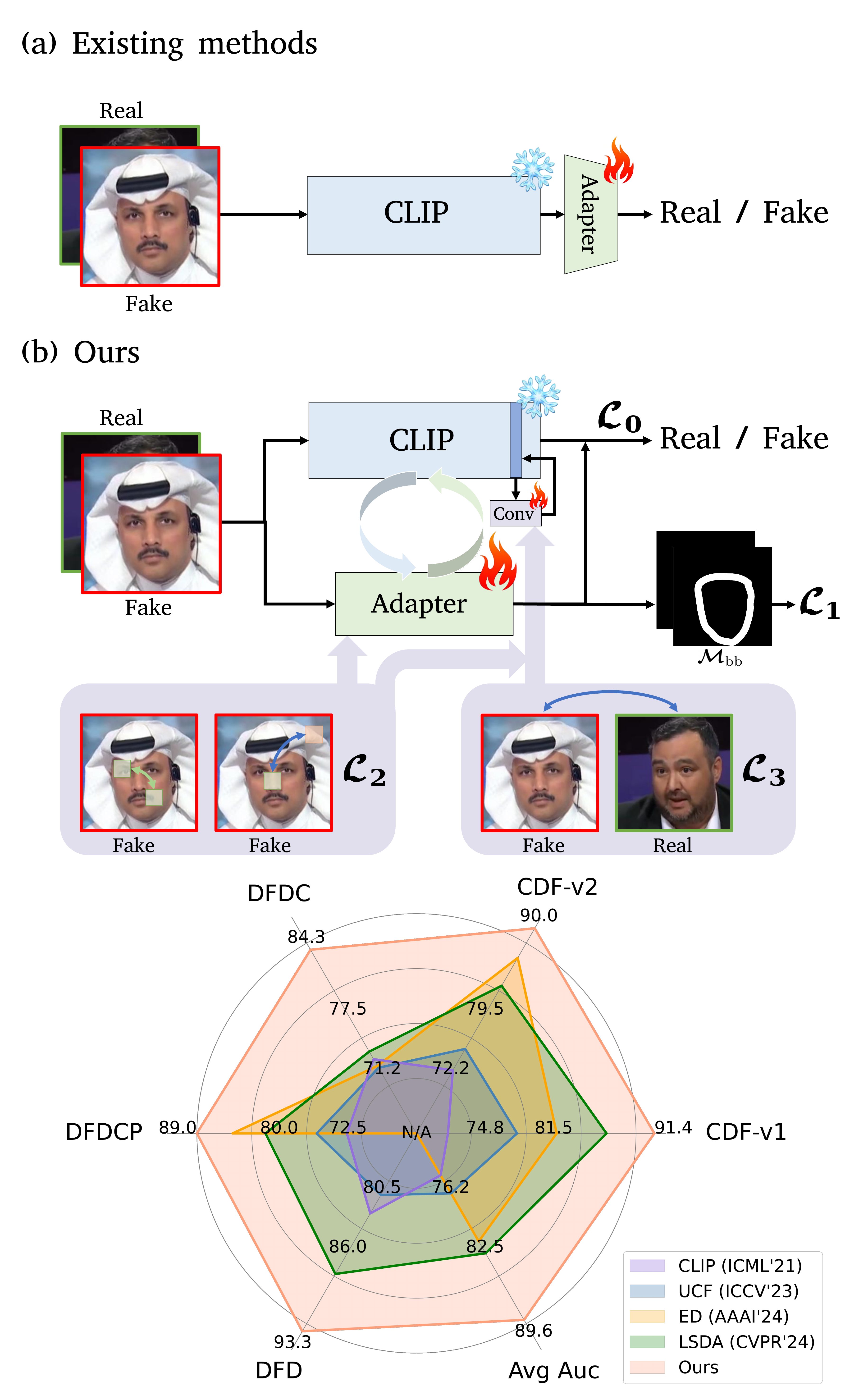}
    \caption{{(a) Existing CLIP-based methods. (b) The proposed Forensics Adapter, achieving the best performance compared with several state-of-the-art methods on CDF-v1~\cite{celeb}, CDF-v2~\cite{celeb}, DFDC~\cite{dfdc}, DFDCP~\cite{dfdcp}, and DFD~\cite{ff++} datasets.}}
    \label{fig:overview}
\end{figure}

Recent face forgery detection methods are typically developed on CNNs, benefiting from the learning capacity of these models. They leverage various forgery clues, including biological signals~\cite{biological_signal, head_poses, heartbeat}, blending artifacts~\cite{ddb,blending_artifacts2,blending_artifacts3}, and frequency signals~\cite{f3net,srm,spsl}. While these methods can achieve favorable and near-perfect performance on standard datasets, their ability to detect unseen forgeries drops significantly. This is because these methods tend to overfit the specific forgeries present in the training data, limiting their generalization across different distributions. With new forgery techniques constantly emerging in real-world scenarios, these detection methods remain highly constrained in practical use. 
To address this issue, several generalizable face forgery detection methods have been proposed, aiming to capture common forgery clues through approaches such as pseudo-fake face augmentations~\cite{SeeABLE,sbi}, and disentanglement learning~\cite{fud,disentanglement}. Nevertheless, given the limited variety of forgery categories, these methods still struggle to acquire sufficient forgery priors. 

CLIP~\cite{clip}, a recent large vision-language model, contains extensive knowledge priors, including that related to forgery, offering promising potential to enhance generalizable face forgery detection. Inspired by this, several methods~\cite{repdfd,ffaa,clipping} have been proposed to utilize CLIP for face forgery detection. They commonly use the frozen CLIP image encoder as a feature extractor, with an optional adapter network appended after CLIP. To learn forgery-related knowledge, the adapter network is trained using general objectives (\eg, cross-entropy) on face forgery datasets (see Fig.~\ref{fig:overview} (a)).

However, this straightforward adaptation has two major limitations: \textbf{(1) Task-agnostic adaptation.} The adapter is designed generically, rather than tailored to 
capture unique task-specific clues in face forgery. \textbf{(2) Lack of interaction between CLIP and the adapter.} Since CLIP is not specifically designed for this task, its forgery-related knowledge is intertwined with a broad range of other knowledge. The adapter, therefore, should interact more directly with CLIP to guide it toward forgery-related knowledge, rather than only refining the final output of CLIP. Moreover, the versatility of CLIP can serve as a powerful source of instruction for training the adapter effectively. 

In this paper, we propose \textbf{Forensics Adapter}, a devoted framework for adapting CLIP to generalizable face forgery detection (see Fig.~\ref{fig:overview} (b)). Unlike previous CLIP-based methods, our method is grounded in the core nature of face forgery, with a task-specific adaptation that focuses on key forgery-related clues. Our motivation stems from the observation that face forgeries often involve localized manipulations, where synthesized content is blended back into the original regions, creating inconsistencies between the manipulated and original regions. These inconsistencies form blending boundaries, which serve as crucial clues specific to face forgery. To capture these subtle traces, we propose a lightweight adapter network alongside CLIP and introduce task-specific objectives to instruct learning of the adapter. Then we describe the interaction strategy between the adapter and CLIP, assisting the learning of the adapter while enhancing the token effectiveness of CLIP. With this adapter, CLIP is transformed into an effective face forgery detector.

While the blended inconsistencies have been utilized in previous methods~\cite{laa-net,sbi}, our method explores the {use of the CLIP model} to provide a new solution to this problem. {{Extensive experiments demonstrate that our method outperforms {state-of-the-art methods} by a large margin across six public datasets, with only $\bm{5.7M}$ trainable parameters.}}
We believe our method can establish a strong baseline for CLIP-based face forgery detection methods.

Furthermore, we describe an extended method, \textbf{Forensics Adapter++}, which further utilizes the multimodal capabilities of CLIP by incorporating textual information. Unlike conventional vision tasks, this incorporation presents unique challenges, as the textual modality should effectively depict subtle forgery traces rather than high-level semantic information. To address this, we design a forgery-aware {prompt} learning strategy with an updated task-specific trace learning scheme, which serves as an auxiliary mechanism boosting the efficacy of visual modality. With negligible extra trainable parameters, Forensics Adapter++ further improves the performance by $\bm{1.3\%}$.

Our contribution can be summarized as follows: 
\begin{itemize}
    \item We bridge the gap between CLIP and {the} face forgery task by introducing a Forensics Adapter that effectively adapts CLIP for generalizable face forgery detection. 

    \item We provide a task-specific adapter design, including both architecture and objectives, and outline an interaction strategy between CLIP and the adapter. 

    \item We further describe Forensics Adapter++, which improves Forensics Adapter {by} incorporating the text modality with specific designs.

    \item The experimental results are promising, demonstrating the potential of our method as a baseline for CLIP-based face forgery detection methods.
\end{itemize}

This work extends our previous study presented at \texttt{CVPR 2025}~\cite{cui}, with the following substantial improvements: (1) We provide more analysis and details regarding the original Forensics Adapter method, comprehensively showing the efficacy of introduced components. (2) We describe Forensics Adapter++, a fresh attempt that incorporates the textual modality to further enhance face forgery detection. To facilitate this, we introduce a forgery-aware prompt learning strategy, which effectively {improves} the capability of visual modality. (3) We conduct additional comparative experiments with {Forensics} Adapter and {Forensics} Adapter++, and more comprehensive ablation studies to analyze the effect of textual modality.
    
    

\section{Related Work}
\label{sec:related-works}

\smallskip
\noindent\textbf{Face Forgery.} Face forgery involves manipulating the local content of the original faces using generative strategies or 3D-based strategies~\cite{ff++,celeb,dfdc,dfdcp,faceswapper}. Its pipeline usually consists of two steps. The first step is to create fake content, \eg, tampered lip or head movement, forged facial identity, etc. The second step is to blend the fake content back to the corresponding position of the original faces while retaining high visual quality. Although the recent face forgery techniques have greatly evolved,  this two-step pipeline always introduces inconsistencies between the forgery and original areas, resulting in a blending boundary. 

Besides face forgery, whole facial image synthesis is another widely studied forensics topic. It involves synthesizing whole images from scratch using GAN~\cite{gan_ori} or Diffusion models~\cite{difussion_ori,diffusion_stable}, based on random noise or certain conditioned inputs. It is important to clarify that this paper focuses on face forgery rather than whole facial image synthesis.


\begin{figure*}[t]
    \centering
    \includegraphics[width=1\linewidth]{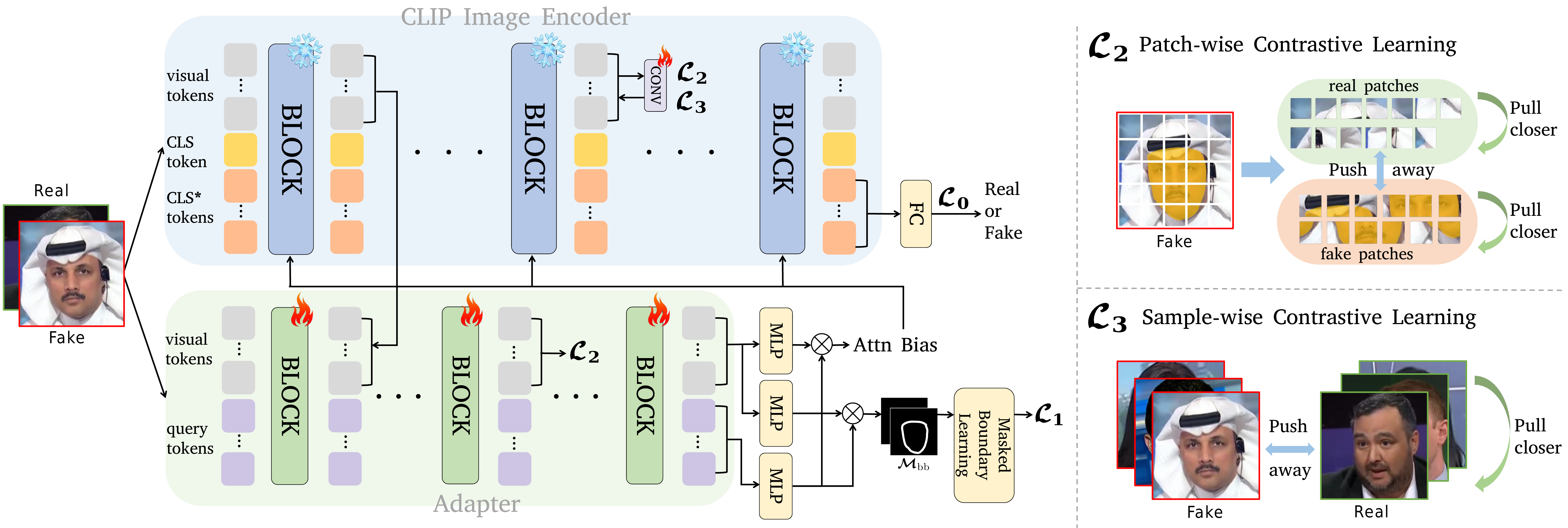}
    \vspace{-0.7cm}
    \caption{Pipeline of the proposed Forensics Adapter in training. The top stream denotes CLIP and the bottom stream corresponds {to the adapter}. See text for details.}
    \label{fig:pipeline}
    \vspace{-0.3cm}
\end{figure*}

\smallskip
\noindent\textbf{Face Forgery Detection.} 
Detecting face forgeries {has become} a significant topic in recent years~\cite{eye_blinking,f3net,laa-net,fud,aunet}. With the great advancement of deep learning, recent methods are typically designed to use Deep Neural Networks (DNNs) to capture the forgery traces, utilizing features ranging from the biological signals (\eg, eye blinking~\cite{eye_blinking}, heartbeat rhythm~\cite{heartbeat}, head pose~\cite{head_poses}, facial action units~\cite{action}), frequency signals (\eg, ~\cite{srm,f3net,spsl}), and auto-learned clues with dedicated training schemes (\eg, \cite{dcl,recce,fud}). However, these methods usually rely on specific training datasets, which can lead to overfitting on known data distributions and hinder their ability to generalize across different datasets. 
To enhance detection generalization, many methods have been proposed to create pseudo-fake faces by simulating blending boundaries in real images~\cite{x-ray,sbi,SeeABLE,laa-net,aunet}.  While these methods improve the generalization to a certain extent, limited data diversity still restricts detectors from learning generic features. {Related studies have also explored forensic robustness under different settings, including post-processed forgery detection~\cite{snis}, compressed Deepfake detection~\cite{famm}, and watermarking-based forensics~\cite{waverecovery,grayscalewm}.}

More recently, several efforts have started adapting CLIP~\cite{clip}, a large vision-language model, for face forgery detection~\cite{repdfd,ffaa,gm-df,udd}. {Many of these approaches have primarily utilized the visual component of CLIP as a feature extractor. } For instance, 
{
UDD~\cite{udd} introduces shuffling and mixing branches to mitigate position and content biases by processing latent tokens with the CLIP visual encoder.
}
GM-DF~\cite{gm-df} attempts to mitigate discrepancies across multiple combined training datasets by leveraging CLIP. 
FCG~\cite{fcg} leverages CLIP to guide the detection of key facial components over temporal sequences.
CLIPping~\cite{clipping} explores the effectiveness of CLIP in combination with recent adaptation methods for universal deepfake detection.
RepDFD~\cite{repdfd} reprograms CLIP by merging universal perturbations {into} the visual input, without modifying the inner parameters of the VLM model.
{Furthermore, some approaches incorporate both the visual and textual components of CLIP. For instance,
VLFFD~\cite{vlffd} generates mixed forgery images with fine-grained prompts and uses them with original data for joint training. 
FFAA~\cite{ffaa} creates a dataset comprising real and forged faces with descriptions and forgery reasoning and implements {face forgery detection} through a multi-answer decision strategy based on CLIP.
}

Regardless of how advanced these methods are, they share a common limitation: they solely use CLIP as a feature extractor, overlooking the unique characteristics of face forgery detection. This results in limited task-specific adaptation and incomplete utilization of CLIP's versatility.

\section{Forensics Adapter}
We describe the Forensics Adapter, designed to capture specific face forgery traces and effectively adapt CLIP into a generalizable face forgery detector.
To ensure practical usability, the adapter should be lightweight. Thus, we employ the tiny vision transformer~\cite{tinyvit} as the foundation of the adapter, incorporating only the first eight layers and positioning it parallel to CLIP to unleash its full capacity, inspired by the methods in ~\cite{faa,side}. Unlike general computer vision tasks, the adapter is instructed to detect the blending boundaries in forged faces, with dedicated objectives (Sec.~\ref{sec:ts}). Meanwhile, we describe an interaction strategy between the adapter and CLIP. On the one hand, absorbing CLIP visual tokens can support the learning process of the adapter, while on the other, the task-specific knowledge gained by the adapter can be transferred back into CLIP (Sec.~\ref{sec:interaction}). 
With these task-specific objectives and interaction strategies, this adapter can effectively instruct CLIP to focus more on forgery traces while preserving its generalizability. The pipeline of our method is shown in Fig.~\ref{fig:pipeline}.


\subsection{Learning Task-specific Traces}
\label{sec:ts}
\smallskip
\noindent\textbf{Input Tokens.} 
{
For the adapter, the input face image is first divided into $16 \times 16$ patches and then converted into visual tokens. Inspired by~\cite{carion2020end,cheng2021per,cheng2022masked,side}, these visual tokens are then concatenated with $N$ learnable query tokens, which are used to learn task-specific knowledge. The combined tokens serve as the input to the adapter.
For CLIP, the same input face image is processed following its standard pipeline, resulting in a set of vision tokens along with a CLS token.
}

\smallskip
\noindent\textbf{Prediction Head.} To instruct the adapter in learning blending boundaries, we design a prediction head in the adapter to predict the blending boundary map. The inputs of this head are query tokens and visual tokens from the adapter. For query tokens, we project {them} to 128 dimensions using {an MLP}, which is denoted as $\bm{Q} \in \mathbb{R}^{N \times 128}$. Similarly, we project visual tokens to 128 dimensions using another {MLP}, which is denoted as $\bm{V}_{\rm bb} \in \mathbb{R}^{h \times w \times 128}$, where $h,w$ {are} $1/16$ of input image size. The inner product of $\bm{Q}$ and $\bm{V}_{\rm bb}$ can generate the blending boundary map as
\begin{equation}
\begin{aligned}
    \bm{\mathcal{M}}_{\rm bb} = {\rm Conv}(\bm{V}_{\rm bb} \bm{Q}^{\top}), 
\end{aligned}
\label{eq:bb}
\end{equation}
where ${\rm Conv}$ denotes a couple of convolution operations that gradually transform the dimension of the output as $\bm{\mathcal{M}}_{\rm bb} \in \mathbb{R}^{h \times w \times 1}$. Each element in $\bm{\mathcal{M}}_{\rm bb}$ represents the probability of being on a blending boundary.
To supervise the prediction and compel CLIP and the adapter to more accurately capture the subtle distinctions between real and fake regions, we generate ground truth blending boundaries for each input and propose three objectives regarding masked blending boundary learning, inter-face contrastive learning, and intra-face patch-wise contrastive learning. The details are elaborated below.

\smallskip
\noindent\textbf{Masked Blending Boundary Learning.}
Following previous works~\cite{x-ray}, we first perform Gaussian blurring on {the} ground truth manipulation mask $\bm{\mathcal{M}}'$ and then generate {the} ground truth blending boundary with $\bm{\mathcal{M}}'_{\rm bb} = 4 \bm{\mathcal{M}}'  (1 - \bm{\mathcal{M}}')$. A straightforward way would be to directly calculate the Mean Squared Error (MSE) between $\bm{\mathcal{M}}'_{\rm bb}$ and $\bm{\mathcal{M}}_{\rm bb}$. However, since the boundary occupies only a small portion of the map, using standard MSE can easily be influenced by the non-boundary areas, leading to suboptimal results. Therefore, we introduce a masking strategy to highlight the effect of the boundary areas. Specifically, we define a binary mask ${\bm{B}}$ and check all $16 \times 16$ patches of $\bm{\mathcal{M}}'_{\rm bb}$. If no boundary exists in a patch, the corresponding elements in $\bm{B}$ are set to $0$. Otherwise, they are set to $1$. 
This objective can be defined as
\begin{equation}
    \mathcal{L}_1 = \mathrm{MSE} (\bm{\mathcal{M}}_{\rm bb} \odot \bm{B},  \bm{\mathcal{M}}'_{\rm bb} ),
\end{equation}
{where $\odot$ denotes element-wise multiplication.}

\smallskip
\noindent\textbf{Patch-wise Contrastive Learning.} 
Note that in a forged face, the representations of the forged region should differ from the authentic regions. Highlighting this difference can assist the model in capturing the task-specific traces. To achieve this, we describe a patch-wise contrastive learning loss performed on the adapter and CLIP.

For the adapter, we extract the intermediate features and perform contrastive learning on them. Each element corresponds to a divided patch, which is labeled as fake if the forgery region covers more than $10\%$ of the corresponding patch, and as real otherwise. Denote $\bm{X} = [\bm{\mathcal{X}}^1, \bm{\mathcal{X}}^0]$ as a feature, where $\bm{\mathcal{X}}^1, \bm{\mathcal{X}}^0$ are two sets containing real and fake elements. The patch-wise contrastive learning can be formulated as
{
\begin{equation}
    \mathcal{L}_2 = - \log  \frac{\exp (\delta(\bm{x}_i,\bm{x}_j){/ \tau})}{\exp (\delta(\bm{x}_i,\bm{x}_j){/ \tau}) + \sum\limits_{\bm{x}_k \in \bm{\mathcal{X}}^*} \exp (\delta(\bm{x}_i,\bm{x}_k){/ \tau})}, 
\label{eq:pw}
\end{equation}
}
where $\bm{x}_i,\bm{x}_j$ are elements from {the} same set, $\bm{\mathcal{X}}^*$ represents the opposite set of $\bm{x}_i$, \ie, $\bm{\mathcal{X}}^* = \bm{\mathcal{X}}^1$ if $\bm{x}_i \in \bm{\mathcal{X}}^0$, and $\bm{\mathcal{X}}^* = \bm{\mathcal{X}}^0$ otherwise. $\tau$ is the temperature parameter and $\delta$ is the cosine similarity.

For CLIP, we apply the same formulation to refine its visual tokens. Note that the parameters of CLIP are frozen and cannot be directly updated. Therefore, we employ another trainable $1 \times 1$ convolution on the visual tokens (\eg, $\bm{X}$) and perform patch-wise contrastive learning on the newly obtained tokens (\eg, $\bm{X}'$). Then we add the enhanced tokens back to CLIP with a factor $\alpha$, that is $\bm{X} = \bm{X} + \alpha \bm{X}'$.

\smallskip
\noindent\textbf{Sample-wise Contrastive Learning.} 
In addition to the patch-wise correlation, there are certain associations between real and fake samples. Since the fake samples in this context may belong to different categories (\eg, DF, NT, etc.), we do not consider the similarity among them. Instead, we only focus on pulling close real samples while increasing the distance between real and fake samples.

Since CLIP focuses more on the global view, we perform sample-wise contrastive learning on it. Denote a training batch as $\bm{D} = [\bm{\mathcal{D}}^1, \bm{\mathcal{D}}^0]$, where $\bm{\mathcal{D}}^1, \bm{\mathcal{D}}^0$ denote the sets with real and fake faces. This learning process can be written as
{
\begin{equation}
    \mathcal{L}_3 = - \log  \frac{\exp (\delta(\bm{X}_i,\bm{X}_j){/ \tau})}{\exp (\delta(\bm{X}_i,\bm{X}_j){/ \tau}) + \sum\limits_{\bm{X}_k \in \bm{\mathcal{D}}^*} \exp (\delta(\bm{X}_i,\bm{X}_k){/ \tau})},
\end{equation}
}
where $\bm{X}_i,\bm{X}_j$ are samples from {the} same set, $\bm{\mathcal{D}}^*$ represents the opposite set of $\bm{X}_i$ as in Eq.~\eqref{eq:pw}. 
Since CLIP is frozen, we use the same strategy in patch-wise contrastive learning to update CLIP. Similarly, we employ another trainable $1 \times 1$ convolution on the visual tokens (\eg, $\bm{X}$) and perform the learning on the newly obtained tokens. Then we add the enhanced tokens back to CLIP with {the} same factor $\alpha$.

\subsection{Interaction between Adapter and CLIP}
\label{sec:interaction}

\smallskip
\noindent\textbf{Absorbing CLIP Visual Tokens.}
The features of CLIP are highly versatile and can greatly support the learning process of the adapter. Therefore, we fuse the features of CLIP into the adapter. Since CLIP and the adapter have different goals, we limit the feature fusion to the shallow layers of the adapter, avoiding interference with the learning of task-specific features. For a standard 24-layer ViT-L/14 model, we fuse features from layers $\{ 1, 8, 16 \}$ to the layers $\{ 1, 2, 3 \}$ of the adapter. Since the feature dimensions of CLIP and the adapter are inconsistent, we employ a $1 \times 1$ convolution to align them. We then add these two features as the fused ones.

\smallskip
\noindent\textbf{Enhancing CLIP Visual Tokens.}
To make CLIP focus more on blending boundary traces, we convey the knowledge of the adapter to CLIP without modifying any parameters. We achieve this by utilizing an attention bias strategy similar to the one proposed in~\cite{side}. Note that CLIP tokens comprise visual tokens and the $\mathrm{[CLS]}$ token, denoted as $[\bm{X}_{\mathrm{vis}},\bm{X}_{\mathrm{[CLS]}}]$. For each layer, we duplicate $N$ copies of $\mathrm{[CLS]}$ token to form a set of independent tokens as $\bm{X}_{\mathrm{[CLS^*]}} = [\bm{X}_{\mathrm{[CLS]}},...,\bm{X}_{\mathrm{[CLS]}}]$. The entire tokens at a layer can be written as $[\bm{X}_{\mathrm{vis}},\bm{X}_{\mathrm{[CLS]}},\bm{X}_{\mathrm{[CLS^*]}}]$. Then we update $\bm{X}_{\mathrm{[CLS^*]}}$ using self-attention based on the output of the adapter and original visual tokens in CLIP, which is defined as
\begin{equation}
    \bm{X}^{(\ell+1)}_{\mathrm{[CLS^*]}} = \mathrm{Softmax}(\bm{Q}^{(\ell)}_{\mathrm{[CLS^*]}} {\bm{K}^{\top}_{\mathrm{vis}}}^{(\ell)} + \bm{\Delta}^{\top}) \bm{V}^{(\ell)}_{\mathrm{vis}},
\end{equation}
where $\ell$ represents the layer number, $\bm{Q}_{\mathrm{[CLS^*]}}$ denotes the query of $\bm{X}_{\mathrm{[CLS^*]}}$, $\bm{K}_{\mathrm{vis}}$ denotes the key of visual tokens $\bm{X}_{\mathrm{vis}}$, and $\bm{V}_{\mathrm{vis}}$ indicates the value of visual tokens $\bm{X}_{\mathrm{vis}}$.

Notably, $\bm{\Delta}$ is the attention bias generated at the end of the adapter, following similar operations in generating the blending boundary map $\bm{\mathcal{M}}_{\rm bb}$ in Eq.~\eqref{eq:bb}. Specifically, we use another {MLP} to project visual tokens to 128 dimensions as $\bm{V}_{\rm ab} \in \mathbb{R}^{h \times w \times 128}$ and generate the attention bias using the inner product of $\bm{Q}$ and $\bm{V}_{\rm ab}$ as
\begin{equation}
\begin{aligned}
    \bm{\Delta} = \bm{V}_{\rm ab} \bm{Q}^{\top}, 
\end{aligned}
\label{eq:ab}
\end{equation}
where $\bm{Q}$ denotes the same query tokens in the prediction head. This attention bias $\bm{\Delta} \in \mathbb{R}^{h \times w \times N}$ adjusts the self-attention calculation in CLIP, serving as the bridge to convey knowledge of the adapter. Finally, we use $\bm{X}_{\mathrm{[CLS^*]}}$ for identification of face authenticity.

\subsection{Overall Objectives}
In the training phase, the main objective is a Cross-entropy loss applied {to} $\bm{X}_{\mathrm{[CLS^*]}}$ in the last layer of CLIP, denoted as $\mathcal{L}_{0}$, to achieve forgery detection. Together with the aforementioned objectives in Sec.~\ref{sec:ts}, the overall objective can be written as
\begin{equation}
    \mathcal{L} = \lambda_0 \mathcal{L}_{0} + \lambda_1 \mathcal{L}_1 + \lambda_2 \mathcal{L}_2 + \lambda_3 \mathcal{L}_3,
\end{equation}
{where $\lambda_{0}$, $\lambda_{1}$, $\lambda_{2}$, and $\lambda_{3}$ are weighting factors to balance each term, and the specific settings are detailed in the experimental section (Sec.~\ref{sec:exp}).}

\begin{figure*}[!ht]
    \centering
    \includegraphics[width=1\linewidth]{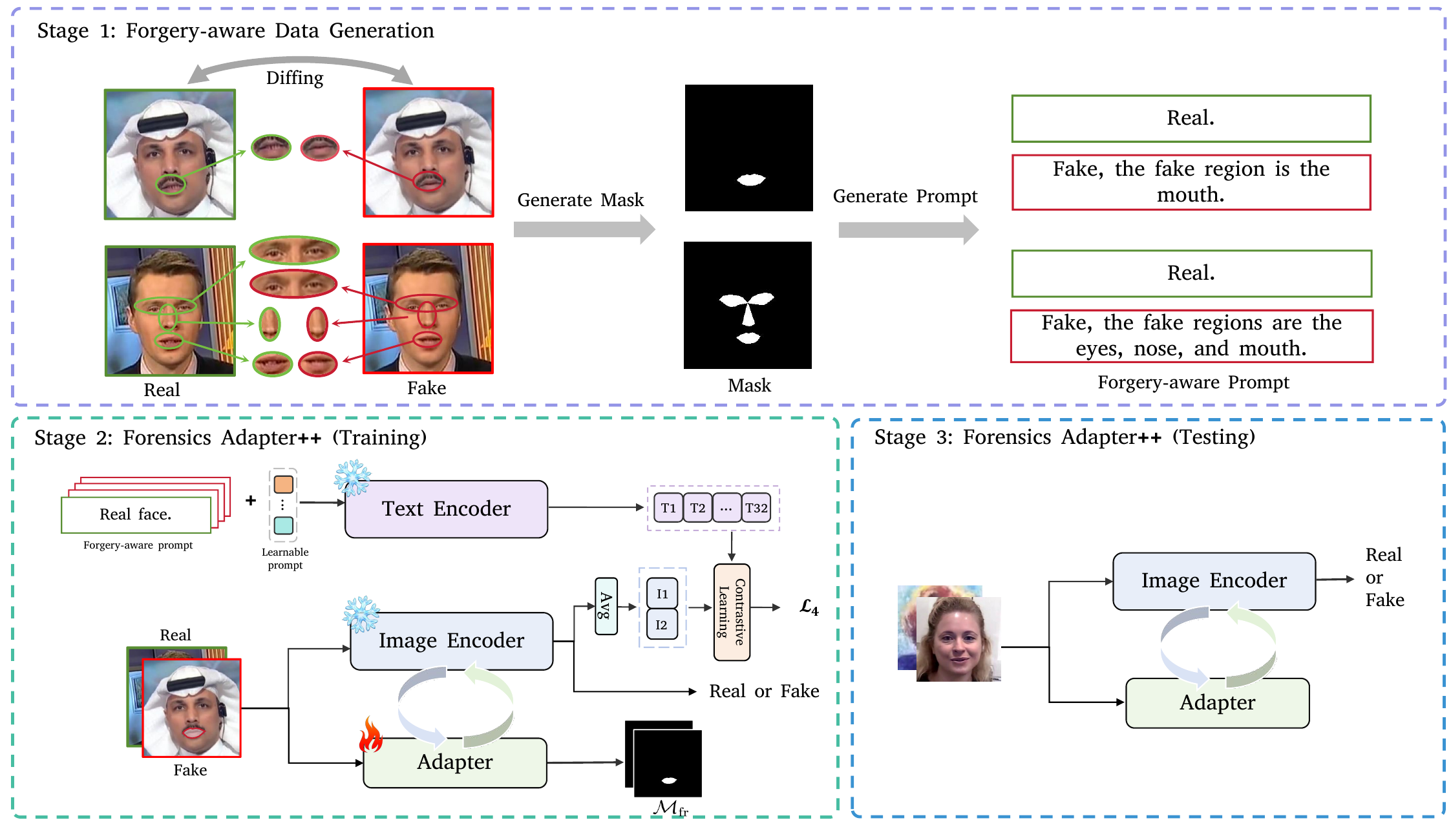}
    \vspace{-0.8cm}
    \caption{
    {Pipeline of Forensics Adapter++. The upper part illustrates the generation of forgery-aware prompts and masks. The lower part shows the training and testing process.
    {Note that the textual modality is only used during training, serving as an auxiliary branch to enhance the visual modality. This diagram highlights the introduction of textual modality while keeping the remaining components consistent with the original Forensics Adapter, except for the updated task-specific traces learning.}}    
    }

    \label{fig:fa++-pipeline}
\end{figure*}

\section{Forensics Adapter++}

The original Forensics Adapter is designed to operate exclusively on the visual modality. Note that CLIP is inherently versatile, supporting both visual and textual modalities. This naturally raises the question: \textit{Can {Forensics Adapter} also benefit from the text modality?}

\subsection{Preliminary Exploration}
Prior works in computer vision have introduced the textual modality through prompt learning approaches~\cite{coop,clip}. However, such a straightforward incorporation of text prompts is not adequate for forensic analysis. To verify this, we froze the text encoder and employed straightforward prompts combined with \textit{real} and \textit{fake} labels to form the final prompts, \eg, {\textit{``This is a real/fake person''}}. The experimental results indicate no performance gains, suggesting that a straightforward incorporation is ineffective for face forensics tasks (see Sec.~\ref{sec:Further-Analysis}).

This highlights the necessity for a tailored design to incorporate textual modality. Upon analysis, we summarize two potential contributing points: \textit{(1) Design of {prompt} learning.} Designing effective prompts for face forensics is non-trivial, as they should reflect forgery-specific information (\eg, subtle manipulation details)  and straightforward prompts that use only the labels \textit{real} and \textit{fake} are ineffective. \textit{(2) Design of incorporation.} Given that visual modality inherently contains richer forgery-related representations, it is more rational to treat textual modality as an auxiliary to visual modality. To fulfill the above points, we describe a forgery-aware prompt learning method with an updated task-specific learning scheme, which introduces more localized evidence to {achieve} better alignment between textual and visual embeddings. The overview of Forensics Adapter++ is illustrated in Fig.~\ref{fig:fa++-pipeline}.


\subsection{Forgery-aware Prompt Learning}

\label{sec:Forgery-aware Prompt}
Forgery-aware prompt learning introduces tailored prompts and dedicated learning objectives. The prompts {consist} of a forgery-aware prompt prefix that emphasizes localized forgery cues, and a learnable prompt suffix that captures implicit contextual forgery cues. The objectives aim to enhance visual representations by aligning them with textual representations.

\smallskip
\noindent\textbf{Forgery-aware Prompt Prefix.} We first partition faces into five different regions based on facial landmarks, denoted as $\{ \mathcal{R}_1,...,\mathcal{R}_5 \}$, corresponding to \textit{eyes, nose, mouth, cheeks}, and \textit{forehead}, respectively. For each fake face, we inspect all regions and determine which regions are manipulated. To do so, we simply calculate the $\ell_2$ pixel-wise color difference between regions in fake and the corresponding real faces, and obtain a forgery score for each region as
\begin{equation}
     s_i =     \frac{1}{\lvert \mathcal{R}_i \rvert} \sum_{p \in \mathcal{R}_i} \mathbb{I} \left ( \ell_2 \left ( \mathcal{I}(p), \mathcal{I}'(p) \right ) \geq \beta \right ),
     \label{eq:si}
\end{equation}
where $\mathcal{I},\mathcal{I}'$ {denote the real and corresponding fake faces}, $p$ denotes the pixel inside $\mathcal{R}_i$, $\mathbb{I}(\cdot)$ is the indicator function that returns 1 if the condition holds, otherwise 0, and the threshold $\beta$ is set to 20. The region $\mathcal{R}_i$ is viewed as forged if $s_i$ exceeds a threshold $\nu$.

Given the forged regions, we enrich the description for fake faces. Besides the real and fake class labels, we additionally introduce which facial parts are forged, such as {\textit{``Fake, the fake regions are the eyes, nose, and mouth.''}}, as shown in Fig.~\ref{fig:fa++-pipeline} (upper). For real faces, we use {\textit{``Real.''}} as the corresponding text description. Overall, we obtain $2^5 = 32$ different types of prompts.

\smallskip
\noindent\textbf{Learnable Prompt Suffix.}
Although the above strategy provides localized evidence, manually designed prompts may still overlook implicit yet effective descriptions. Therefore, inspired by CoOp~\cite{coop}, we concatenate $K$ learnable context {tokens} to make prompts adaptive to objectives. The concatenated prompts serve as the final prompts for the text encoder.

\smallskip
\noindent\textbf{Incorporation Objectives.}
The textual modality should be introduced as an auxiliary to enhance the visual representations. To achieve this, we conduct visual-textual alignment using the cosine similarity between the encoded textual features $\bm{X}_{\mathrm{tex}}$ and the token $\bm{X}_{\mathrm{[CLS^*]}}$ from the CLIP visual encoder. The cross-entropy loss is then utilized for this purpose:
\begin{equation}
\begin{aligned}
& \mathcal{L}_4 = - \log \left (
\frac{
\exp(\mathrm{cos}(\mathrm{Avg}(\bm{X}_{\mathrm{[CLS^*]}}),\bm{X}^{c}_{\mathrm{tex}}))
}{
\sum_{i} \exp(\mathrm{cos}(\mathrm{Avg}(\bm{X}_{\mathrm{[CLS^*]}}),\bm{X}^{i}_{\mathrm{tex}}))
}
\right ),
\end{aligned}
\end{equation}
where $\mathrm{Avg}$ means average pooling to align the dimension of features from the textual and visual {branches}, $c$ denotes the prompt type of input face.
\textit{It is worth noting that this alignment process is used only during training. At the inference stage, we still rely on $\bm{X}_{\mathrm{[CLS^*]}}$ for face forgery detection.}

\subsection{Updated Task-specific Traces Learning}

{In the original Forensics Adapter, the blending boundary is learned as an effective indicator specific to face forgery. By incorporating textual modality, this learning process is updated to better exploit the localized evidence described in Sec.~\ref{sec:Forgery-aware Prompt}. Specifically, we utilize the forged-region mask instead of the blending-boundary mask (see Fig.~\ref{fig:fa++-pipeline} (left)). Unlike the sparse blending-boundary supervision, which is easily dominated by non-boundary regions, this region-level supervision is spatially denser and covers a larger manipulated area, while still implicitly reflecting the blending boundary. Therefore, the masking strategy in the original Masked Blending Boundary Learning is no longer required, and the objective can be updated as}

\begin{equation}
    \mathcal{L}_1 = \mathrm{MSE} (\bm{\mathcal{M}}_{\rm fr},  \bm{\mathcal{M}}'_{\rm fr} ),
\end{equation}
where $\bm{\mathcal{M}}'_{\rm fr}$ denotes the mask of forged regions, and $\bm{\mathcal{M}}_{\rm fr}$ is the predicted result equivalent to $\bm{\mathcal{M}}_{\rm bb}$ in Eq.~\eqref{eq:bb}.
Together with the previously defined losses, the final loss is expressed as:
\begin{equation}
    \mathcal{L} = \lambda_0 \mathcal{L}_{0} + \lambda_1 \mathcal{L}_1 + \lambda_2 \mathcal{L}_2 + \lambda_3 \mathcal{L}_3 + \lambda_4 \mathcal{L}_4,
\end{equation}
{where $\lambda_{0}$, $\lambda_{1}$, $\lambda_{2}$, $\lambda_{3}$, and $\lambda_{4}$ are weighting factors, and their specific settings are detailed in Sec.~\ref{sec:exp}.}

\section{Experiments}

\subsection{Experimental Settings}
\label{sec:exp}

\smallskip
\noindent\textbf{Datasets.}
Following previous methods~\cite{deepfakebench,cfm,diffusionfake}, we utilize the following seven publicly available datasets: FaceForensics++ (FF++)~\cite{ff++}, Deepfake Detection Challenge (DFDC)~\cite{dfdc}, preview version of DFDC (DFDCP)~\cite{dfdcp}, two versions of CelebDF (CDF-v1, CDF-v2)~\cite{celeb}, WildDeepfake (WDF)~\cite{wdf} and DeepfakeDetection (DFD)~\cite{ff++}. To validate the generalizability, our {methods}, Forensics Adapter and Forensics Adapter++, {are} trained on the c23 compression version of FF++ and tested on other datasets.





\begin{table*}[!htbp]
\caption{Cross-dataset evaluation results \textbf{(Frame-level AUC)}. All methods are trained on FF++ and evaluated on other datasets. The \textbf{best} results are indicated in bold and the \underline{second-best} results are underlined.}
\centering
\small
\vspace{-0.3cm}
\begin{tabular}{c|c|cccccc} 
\toprule 
Method  & Venue & CDF-v1 & CDF-v2&  DFDC &  DFDCP & DFD & WDF\\
\midrule 
Xception~ \cite{ff++} &ICCV'19 &0.779 & 0.737 &0.708  & 0.737 & 0.816 & -\\
EfficientB4 \cite{efficientnet} &ICML'19& 0.791 & 0.749 &0.696  &0.728 & 0.815 & -  \\
F3Net \cite{f3net} &AAAI'20  & 0.777 & 0.735 & 0.702  & 0.735 & 0.798 & - \\
X-ray \cite{x-ray} &CVPR'20  & 0.709 & 0.679 & 0.633  & 0.694 & 0.766 & -\\
FFD \cite{ffd} &CVPR'20 & 0.784 & 0.744 & 0.703  &0.743 & 0.802 &  -\\
SPSL \cite{spsl} &CVPR'21  & 0.815 & 0.765 & 0.704  & 0.741 & 0.812 & -\\
SRM \cite{srm} &CVPR'21  & 0.793 & 0.755 & 0.700  & 0.741 & 0.812 & - \\
Recce \cite{recce}& CVPR'22  & 0.768 & 0.732 & 0.713  & 0.734  & 0.812  & - \\
SBI \cite{sbi} &CVPR'22   & -& 0.813 & -  & 0.799  & 0.774   & 0.672 \\ 
UCF \cite{ucf} &ICCV'23  & 0.779&0.753 & 0.719 & 0.759  & 0.807 & - \\
ED \cite{ba}&AAAI'24  & 0.818&{0.864} & 0.721 & {0.851}  & - & -  \\ 
CFM \cite{cfm} &TIFS'24  & -& 0.828 & - & 0.758  & {0.915} & 0.784 \\ 
FoCus \cite{focus} &TIFS'24 &-&0.720 &0.669 & 0.778& -&0.722 \\
LSDA \cite{lsda} &CVPR'24  &0.867& 0.830 & {0.736} & 0.815  & 0.880 & - \\
DiffusionFake \cite{diffusionfake}& NeurIPS'24  & - &0.805& -  & 0.810  & 0.904 &0.801 \\ 
UDD \cite{udd}&AAAI'25  &- & 0.869 & 0.758 & 0.856  & 0.910   & - \\ 
\midrule 
\rowcolor{hl} {ForAda} (ours) &CVPR'25& \underline{0.914} &\underline{0.900} & \underline{0.843} & \underline{0.890} & \textbf{0.933}&  \underline{0.803} \\
\rowcolor{hl} {ForAda++}  (ours) &- & \textbf{0.926} & \textbf{0.903} & \textbf{0.863} & \textbf{0.900} & \underline{0.922}&  \textbf{0.850}\\
\bottomrule 
\end{tabular}
\vspace{-0.3cm}
\label{tab:frame} 
\end{table*}

\begin{table*}[!t]
\caption{Cross-dataset evaluation results \textbf{(Video-level AUC)}. The top section presents video-based methods and the bottom section corresponds to frame-based methods. } 
\centering
\small
\vspace{-0.3cm}
\begin{tabular}{c|c|ccccc} 
\toprule 
Method & Venue  & CDF-v2  &  DFDC &  DFDCP & DFD &  WDF  \\
\midrule 
TALL \cite{tall} &ICCV'23  & 0.908 & 0.768 & - & - & -\\  
SeeABLE \cite{SeeABLE}& ICCV'23  & 0.873 &  0.759 & 0.863 & - &-  \\ 
IID \cite{iid} &CVPR'23  &0.838 & - & 0.812 &- &-    \\ 
TALL++ \cite{tall++} &IJCV'24  &  0.920 &  {0.785} & -&-&  \\ 
SAM \cite{sam} &CVPR'24  & 0.890 & - & - & 0.961& -   \\ 

\midrule 
SBI \cite{sbi} &CVPR'22   & 0.932 & 0.724 & 0.862 & -& -  \\ 
CADDM \cite{caddm}& CVPR'23  & 0.939 &  0.739 &- &- &-   \\
SFDG \cite{sfgd}& CVPR'23  & 0.758 &  0.736 & - &0.880 & 0.692   \\
LAA-NET \cite{laa-net}& CVPR'24  &\underline{0.954} & - & {0.869} &0.800&-   \\ 
LSDA \cite{lsda} &CVPR'24  &0.898 &0.735 & 0.812& 0.956& 0.756  \\
FreqBlender \cite{freqblender}& NeurIPS'24  & 0.746 &-& 0.876  & -  & -  \\ 
CFM \cite{cfm} &TIFS'24 &0.897 & 0.706 & 0.802 &0.952 & \underline{0.823}  \\ 


\midrule 
\rowcolor{hl}{ForAda}  & CVPR'25 & \textbf{0.957} & \underline{0.872} & \underline{0.929} & \textbf{0.972} & 0.805  \\
\rowcolor{hl}{ForAda++}  & -  & 0.940 & \textbf{0.887}  & \textbf{0.947} &\underline{0.965} & \textbf{0.867}\\

\bottomrule 
\end{tabular}
\vspace{-0.3cm}
\label{tab:video} 
\end{table*}

\smallskip
\noindent\textbf{Implementation Details.}
{Our method is implemented using PyTorch 2.3.0~\cite{pytorch} and trained on a single NVIDIA RTX 3090 GPU.}
{The adapter is configured with a patch size of 16 and input images resized to 256×256 pixels, without using pre-trained weights. {The CLIP architecture is ViT-L/14, with a patch size of 14 and input images resized to 224×224 pixels, initialized with pre-trained weights.}}
The number of learnable query tokens $N$ is set to 128. For the adapter, we use the features from intermediate layers 4, 5, and 6 for patch-wise contrastive learning. For CLIP, we add a trainable $1 \times 1$ convolution to its 13th layer to support both sample-wise contrastive learning and patch-wise contrastive learning, with the factor $\alpha$ set to 0.05. For overall objectives, $\lambda_{2}$ is different on the adapter and CLIP: $\lambda_{2} = 20$ on the adapter and $\lambda_{2} = 10$ on CLIP. Other factors are set to $\lambda_0 = 10, \lambda_1=200, \lambda_3=10$.
The entire model was trained on the FF++ c23 training set, with a batch size of 16, using the Adam optimizer~\cite{adam} with a learning rate of 0.0002 and weight decay of 0.0005. We follow the configuration of DeepfakeBench~\cite{deepfakebench}: For each video, 32 frames were extracted for training or testing, and RetinaFace~\cite{RetinaFace} was used for face extraction.

In Forensics Adapter++, we adopt the same visual configuration as in the original Forensics Adapter. {Following~\cite{clip}, we use the standard 12-layer Transformer architecture as the text encoder, initialized with pre-trained weights.} The number of learnable context tokens is set to 8 and the threshold $\nu$ is fixed at 0.15. We set the loss weight $\lambda_4 = 1.5$, while keeping all other loss weights $\lambda_{0 \sim 3}$ unchanged.

\subsection{Results}
\smallskip

\noindent\textbf{Compared with State-of-the-art Methods.} 
For a comprehensive comparison, we compare our method with other state-of-the-art methods under both frame-level and video-level evaluations.

Table~\ref{tab:frame} shows the cross-dataset evaluation results in frame-level AUC. In this scenario, all methods are trained on the FF++ c23 set and evaluated across different datasets.
Note that the results of CFM~\cite{cfm}, ED~\cite{ba}, FoCus~\cite{focus}, LSDA~\cite{lsda}, DiffusionFake~\cite{diffusionfake} and UDD~\cite{udd} are taken from their original papers. Since SBI~\cite{sbi} does not report its frame-level results, we use the reproduced results from CFM. For all remaining methods, we take the results from LSDA~\cite{lsda}.

Our Forensics Adapter (denoted as ForAda) outperforms existing state-of-the-art methods on six datasets -- CDF-v1, CDF-v2, DFDC, DFDCP, DFD and WDF -- by margins of $\bm{4.7\%}$, $\bm{3.1\%}$, $\bm{8.5\%}$, $\bm{3.4\%}$, $\bm{1.8\%}$, $\bm{0.2\%}$, respectively, highlighting the effectiveness of the proposed adaptation strategies.
{Moreover, Forensics Adapter++ (denoted as ForAda++) improves the average performance of ForAda by $\bm{1.3\%}$. Specifically, ForAda++ yields gains of $\bm{1.2\%}$, $\bm{0.3\%}$, $\bm{2.0\%}$, $\bm{1.0\%}$, and $\bm{4.7\%}$ on CDF-v1, CDF-v2, DFDC, DFDCP, and WDF, respectively, while exhibiting a $\bm{1.1\%}$ decrease on DFD. We conjecture that, for relatively complex datasets, visual forgery cues tend to be more subtle, and incorporating textual modality can more effectively assist in localizing manipulated regions and uncovering forgery-related clues. In contrast, on relatively simpler datasets, visual information alone is already sufficiently discriminative, thereby limiting the additional benefit of textual modality. Overall, ForAda++ still achieves superior average performance over all six datasets, demonstrating the overall benefit of incorporating textual modality.
}

\begin{table}[!t]
\caption{Cross-dataset evaluation results \textbf{(Video-level AUC)}. }  
\small%
\centering
\vspace{-0.3cm}
\resizebox{\linewidth}{!}{
\begin{tabular}{c|c|ccc} 
\toprule 
Method  & {Venue} & CDF-v2  &  DFDC &  DFD  \\
\midrule 
{Vanilla CLIP}~\cite{clip} & ICML'21 & 0.777 & 0.742 & 0.834 \\
\midrule 
FFAA \cite{ffaa}  & arXiv'24 & - & 0.740 & 0.920 \\

GM-DF \cite{gm-df} & arXiv'24 & 0.832 & 0.772 & -   \\  
CLIPping \cite{clipping}  & ICMR'24 & - & 0.719 & 0.866   \\
RepDFD \cite{repdfd}  & AAAI'25 & 0.899 & 0.810 & -   \\  

VLFFD \cite{vlffd}  & CVPR'25 & 0.848 & - & {0.948}   \\ 
FCG \cite{fcg} & CVPR'25 & \underline{0.950} & 0.818 & -   \\

\midrule 
\rowcolor{hl}{ForAda}  & CVPR'25 & \textbf{0.957} & \underline{0.872} & \textbf{0.972} \\

\rowcolor{hl}{ForAda++}  & -  & {0.940} & \textbf{0.887} & \underline{0.965}  \\
\bottomrule 
\end{tabular}
}
\vspace{-0.3cm}
\label{tab:clip} 
\end{table}

Table~\ref{tab:video} shows the cross-dataset evaluation results in video-level AUC.
Following previous works~\cite{cfm}, the evaluation is performed on CDF-v2, DFDC, DFDCP, DFD and WDF datasets.
The top section presents video-based methods and the bottom section corresponds to frame-based methods\footnote{Since several frame-based methods only report either frame-level AUC or video-level AUC, the included methods in Table~\ref{tab:frame} and Table~\ref{tab:video} (bottom section) are not fully consistent.}. Note that SeeABLE~\cite{SeeABLE}, SBI~\cite{sbi} and LAA-NET~\cite{laa-net} are trained on the custom dataset, while others are trained on FF++ c23 set. 
{Our ForAda achieves substantial performance improvements on most datasets, improving AUC by $\bm{0.3\%}$, $\bm{8.7\%}$, $\bm{5.3\%}$, and $\bm{1.1\%}$ on CDF-v2, DFDC, DFDCP, and DFD, respectively. These results demonstrate the effectiveness of our method in generalizable face forgery detection. Furthermore, while ForAda++ shows slight performance decreases on certain datasets, it still achieves an additional average improvement of $\bm{1.4\%}$ over ForAda. This trend is largely consistent with the observations in the frame-level setting, suggesting that incorporating textual modality tends to be more beneficial for complex forgery scenarios.}

\smallskip
\noindent\textbf{Compared with CLIP-based Methods.}
To fully demonstrate the effectiveness of the adapter, we compare our method with CLIP-based methods. 
Table~\ref{tab:clip} shows the comparison results of our method and other CLIP-based methods. Note that VLFFD~\cite{vlffd}, CLIPping~\cite{clipping}, and FFAA~\cite{ffaa} are trained on the custom dataset, while the others are trained on FF++ c23 set. Except for CLIPping, whose results are taken from FFAA, all other results are taken directly from the original papers. Vanilla CLIP refers to the frozen ViT-L/14 model with an additional FC head. We can see that our method notably outperforms others on all datasets, improving performance by $\bm{0.7\%}$, $\bm{6.9\%}$, and $\bm{2.4\%}$ on CDF-v2, DFDC, and DFD, respectively. It further demonstrates the superiority of the proposed adapter.

\subsection{Ablation Studies on Forensics Adapter}
\label{sec:Analysis}

\begin{table*}[!ht]
\caption{Ablation study on the effect of three losses. Evaluation metrics are the frame-level AUC, AP, and EER, respectively.}
\vspace{-0.3cm}
\setlength{\tabcolsep}{4pt} 
\small%
\centering
\begin{tabular}{c|c|c|ccc|ccc|ccc|ccc|ccc}
\toprule
 \multirow{2}{*}{$\mathcal{L}_1$} & \multirow{2}{*}{$\mathcal{L}_2$} & \multirow{2}{*}{$\mathcal{L}_3$} & \multicolumn{3}{c|}{CDF-v1} & \multicolumn{3}{c|}{CDF-v2} & \multicolumn{3}{c|}{DFDC} & \multicolumn{3}{c|}{DFDCP} & \multicolumn{3}{c}{DFD} \\ 
\cmidrule(lr){4-6} \cmidrule(lr){7-9} \cmidrule(lr){10-12} \cmidrule(lr){13-15} \cmidrule(lr){16-18} 
& & & AUC & AP & EER & AUC & AP & EER & AUC & AP & EER & AUC & AP & EER & AUC & AP & EER  \\
\midrule
 $\times$ & $\checkmark$& $\checkmark$ & 0.887 & 0.925 & 19.3 & 0.860 & 0.926 & 22.3 & 0.825 & 0.857 & 25.5 & 0.848 & 0.921 & 23.5& 0.918 &  0.990 & 15.1 \\  

$\checkmark$ & $\times$& $\checkmark$ & 0.818 &0.869 & 26.1 & 0.822 &  0.885 & 25.8 &0.790& 0.805 & 27.8 & 0.826 &0.896& 23.9  & 0.902 &0.987& 16.9 \\ 

$\checkmark$ & $\checkmark$& $\times$ & 0.904 &0.934 &18.1 &0.899 & 0.941 & 18.8 & 0.840 & 0.871& 24.5 & \textbf{0.893} &\textbf{0.943} & \textbf{19.0} &  0.920 &0.990 &15.2\\
$\checkmark$ & $\checkmark$ & $\checkmark$& \textbf{0.914} & \textbf{0.940} & \textbf{15.8} & \textbf{0.900} & \textbf{0.945} & \textbf{18.2} & \textbf{0.843} & \textbf{0.873} & \textbf{23.9} & 0.890 & 0.942 & 19.7 & \textbf{0.933} & \textbf{0.992}& \textbf{13.4} \\ 
\bottomrule
\end{tabular}

\label{tab:loss}
\end{table*}

\begin{table*}[!ht]
\caption{{Ablation study on where to apply task-specific contrastive learning.}}
\vspace{-0.3cm}
\setlength{\tabcolsep}{4pt} 
\small%
\centering
\begin{tabular}{c|c|ccc|ccc|ccc|ccc|ccc}
\toprule
\multirow{2}{*}{ADP} & \multirow{2}{*}{CLIP} & \multicolumn{3}{c|}{CDF-v1} & \multicolumn{3}{c|}{CDF-v2} & \multicolumn{3}{c|}{DFDC} & \multicolumn{3}{c|}{DFDCP} & \multicolumn{3}{c}{DFD} \\ 
\cmidrule(lr){3-5} \cmidrule(lr){6-8} \cmidrule(lr){9-11} \cmidrule(lr){12-14} \cmidrule(lr){15-17}
& & AUC & AP & EER & AUC & AP & EER & AUC & AP & EER & AUC & AP & EER & AUC & AP & EER \\
\midrule
$\times$ & $\times$ & 0.814 & 0.895 & 24.9 & 0.789 & 0.880 & 28.7 &  0.744 & 0.778 & 32.3 & 0.748 & 0.868 & 33.0 & 0.854 & 0.982 & 22.4 \\  

$\checkmark$ & $\times$ & 0.845 & 0.912 & 24.0 & 0.822 & 0.899 & 25.8 & 0.778 &  0.817 & 29.8 & 0.809 &  0.893 & 28.3 & 0.899 & 0.987 & 18.5\\  

$\times$ & $\checkmark$ & 0.881 &0.926 & 20.3 & 0.875 & 0.923 & 21.4& 0.817 &0.840 & 26.2 & 0.833 & 0.894 & 24.0 & 0.903 & 0.988& 17.3\\ 

$\checkmark$ & $\checkmark$ & \textbf{0.914} & \textbf{0.940} & \textbf{15.8} & \textbf{0.900} & \textbf{0.945} & \textbf{18.2} & \textbf{0.843} & \textbf{0.873} & \textbf{23.9} & \textbf{0.890} & \textbf{0.942} & \textbf{19.7} & \textbf{0.933} & \textbf{0.992}& \textbf{13.4} \\ 

\bottomrule
\end{tabular}
\label{tab:task}
\end{table*}

\begin{table*}[!t]
\caption{Ablation study on the number of layers used in adapter.}
\vspace{-0.3cm}
\setlength{\tabcolsep}{4pt} 
\small%
\centering
\begin{tabular}{c|ccc|ccc|ccc|ccc|ccc}
\toprule
Adapter  & \multicolumn{3}{c|}{CDF-v1} & \multicolumn{3}{c|}{CDF-v2} & \multicolumn{3}{c|}{DFDC} & \multicolumn{3}{c|}{DFDCP} & \multicolumn{3}{c}{DFD} \\ 
\cmidrule(lr){2-4} \cmidrule(lr){5-7} \cmidrule(lr){8-10} \cmidrule(lr){11-13} \cmidrule(lr){14-16}
layers &  AUC & AP & EER & AUC & AP & EER & AUC & AP & EER & AUC & AP & EER & AUC & AP & EER \\
\midrule

$\bm{6}$ &  0.890 &  0.928 & 19.0& 0.878 &0.928 &20.8 & 0.836 & 0.864 &24.8 & 0.869&0.923 & 21.6 & \textbf{0.936}  &\textbf{0.993} &\textbf{13.2}\\  

$\bm{8}$& \textbf{0.914} & \textbf{0.940} & \textbf{15.8} & 0.900 & \textbf{0.945} & 18.2 & {0.843} & {0.873} & {23.9} & \textbf{0.890} & \textbf{0.942} & \textbf{19.7} & {0.933} & {0.992}& {13.4} \\  

$\bm{10}$ &  0.900 &0.930& 17.0 &\textbf{0.901} & 0.936 & \textbf{18.0} & \textbf{0.848} &   \textbf{0.883} & \textbf{23.0}& 0.878 &0.931& 20.4 &  0.910 &  0.989 & 16.4 \\

$\bm{12}$ &  0.877 &0.907& 21.1 & 0.888& 0.931 & 20.0 &0.825 & 0.857 & 26.0 &  0.870 &0.929 &21.3 & 0.909 &  0.988 & 16.5 \\ 
\bottomrule
\end{tabular}

\label{tab:adapter_layer}
\end{table*}

\smallskip
\noindent\textbf{Effect of Losses.}
This part studies the effect of each loss ($\mathcal{L}_1$, $\mathcal{L}_2$, $\mathcal{L}_3$). We employ Area Under the Curve (AUC), Average Precision (AP), and Equal Error Rate (EER) for evaluation. The results are shown in Table~\ref{tab:loss}. It can be seen that using all losses achieves the best performance.
Notably, $\mathcal{L}_2$ seemingly has the largest impact on performance, improving AUC and AP by $6.4\%$ and $5.0\%$, respectively, while reducing the EER by an average of $5.9\%$.

\smallskip
\noindent\textbf{{Where to Apply Task-specific Contrastive Learning.}}
Theoretically, patch-wise contrastive learning and sample-wise contrastive learning can be used on both CLIP and {the} adapter. Thus, we study where the learning should {be} added. The results, shown in Table~\ref{tab:task}, highlight the importance of the proposed task-specific learning strategies. By using the learning on the adapter (ADP) side, {we can improve performance by $4.1\%$ and $2.1\%$ in AUC and AP, respectively, while reducing EER by $3.0\%$ on average.} By applying the learning solely on the CLIP side, {we can achieve a performance increase of around $7.2\%$ and $3.3\%$ in AUC and AP, respectively, while lowering EER by $6.5\%$ on average.} When adding the learning strategy to both CLIP and the adapter, our method achieves the best performance.

\smallskip
\noindent\textbf{Number of Layers Used in Adapter.}
In the main experiment, we use the first 8 layers of Tiny-ViT as the adapter. To comprehensively investigate the effect of the adapter size, we conduct additional experiments using adapters with various numbers of layers (6, 8, 10, and 12). The results are presented in Table~\ref{tab:adapter_layer}. We can observe that using 8 layers yields better performance than 6 layers. Nevertheless, increasing the number of layers beyond 8 does not lead to further improvements. This suggests that 8 layers are sufficient to capture comprehensive forgery traces, and adding more layers does not necessarily enhance performance.

\smallskip
\noindent\textbf{Area Threshold for Determining Fake Patches.}
This part studies the optimal threshold for determining a patch as fake based on the proportion of fake content within it, measured by its overlap with the ground truth mask. We experiment with thresholds of 10\%, 50\%, 90\%, and 100\%. {A threshold of} 10\% is used in {the} main experiment, and 100\% means a patch is viewed as fake if it has {full} overlap. The results shown {in Table~\ref{tab:fake_propor}} reveal a performance decline {as the threshold increases}. This is because higher thresholds exclude patches with partial forgeries, which still carry meaningful forgery cues.

\begin{table}[!ht]
\caption{Ablation study on the area threshold for determining fake patches.}
\vspace{-0.3cm}
\setlength{\tabcolsep}{4pt} 
\small
\centering
\begin{tabular}{c|ccc|ccc}
\toprule
Area  & \multicolumn{3}{c|}{CDF-v2} & \multicolumn{3}{c}{DFDC}  \\ 
\cmidrule(lr){2-4} \cmidrule(lr){5-7}
threshold &  AUC & AP & EER & AUC & AP & EER  \\
\midrule
10\% & \textbf{0.900} & \textbf{0.945} & \textbf{18.2} & \textbf{0.843} & \textbf{0.873} & \textbf{23.9}\\ 
50\% & 0.876 & 0.921 & 20.1 & 0.810 & 0.828 &  26.0 \\
90\% &  0.854 & 0.913 & 22.9 & 0.795 &  0.823  &28.3\\
100\% &  0.855 & 0.905 & 21.6 & 0.782 & 0.812 & 28.3\\
\bottomrule
\end{tabular}
\vspace{-0.3cm}
\label{tab:fake_propor}
\end{table}

\smallskip
\noindent\textbf{Effect of Masked Blending Boundary Learning.}
In this section, we investigate the effectiveness of our Masked Blending Boundary Learning. Specifically, we discard the use of the binary mask ${\bm{B}}$ and directly apply MSE loss on $\bm{\mathcal{M}}'_{\rm bb}$ and $\bm{\mathcal{M}}_{\rm bb}$. The comparison results are presented in Table ~\ref{tab:withB}, demonstrating that the use of ${\bm{B}}$ {is} still necessary.

\begin{table}[!ht]
\caption{Ablation study on the effect of Masked Blending Boundary Learning.}
\vspace{-0.3cm}
\setlength{\tabcolsep}{4pt} 
\small
\centering
\begin{tabular}{c|ccc|ccc}
\toprule
\multirow{2}{*}{Setting}  & \multicolumn{3}{c|}{CDF-v2} & \multicolumn{3}{c}{DFDC}  \\ 
\cmidrule(lr){2-4} \cmidrule(lr){5-7}
&  AUC & AP & EER & AUC & AP & EER  \\
\midrule
w/o ${\bm{B}}$ & 0.893 & 0.941 & 18.7 & 0.837 & 0.867 &24.7\\ 
w/ ${\bm{B}}$ &\textbf{0.900} & \textbf{0.945} & \textbf{18.2} & \textbf{0.843} & \textbf{0.873} & \textbf{23.9}  \\
\bottomrule
\end{tabular}
\vspace{-0.3cm}
\label{tab:withB}
\end{table}

\subsection{Ablation Studies on Forensics Adapter++}



\smallskip
\noindent{\textbf{Effect of Textual Modality.}
This section examines the contribution of textual modality in ForAda++. As shown in Table~\ref{tab:tt}, \emph{w/o text} denotes the variant without textual modality, \emph{w/ text (shared prompts)} employs the unified prompts “A photo of a real face.” and “A photo of a fake face.” for real and fake samples, respectively, while \emph{w/ text (region-aware prompts)} uses the region-aware prompt design adopted in the main experiments. Compared with \emph{w/o text}, introducing shared prompts leads to only marginal performance changes, suggesting that the auxiliary text branch preserves model stability but that such simple prompts provide limited semantic guidance. In contrast, region-aware prompts consistently outperform the other variants on most challenging datasets, indicating that finer-grained textual semantics better guide the model’s attention to localized forged regions.
}

\begin{table}[!ht]
\caption{{Ablation Study on the effect of textual modality.}}
\vspace{-0.3cm}
\setlength{\tabcolsep}{4pt} 
\small
\centering
\resizebox{\linewidth}{!}{
\begin{tabular}{c|ccc|ccc}
\toprule
\multirow{2}{*}{Setting}  & \multicolumn{3}{c|}{CDF-v2} & \multicolumn{3}{c}{DFDC}  \\ 
\cmidrule(lr){2-4} \cmidrule(lr){5-7}
&  AUC & AP & EER & AUC & AP & EER  \\
\midrule
w/o text  & 0.888 & 0.936 & 19.2 & 0.838 & 0.869 &23.9\\ 
w/ text (shared prompts) & 0.893 & 0.938 & 18.1 & 0.837 & 0.871 &24.0\\ 
w/ text (region-aware prompts) &\textbf{0.903} & \textbf{0.943} & \textbf{17.8} & \textbf{0.863} & \textbf{0.884} & \textbf{21.9}  \\
\bottomrule
\end{tabular}
}

\vspace{-0.3cm}
\label{tab:tt}
\end{table}

\smallskip
\noindent{\textbf{Effect of the Masking Strategy.}
This section examines whether the masking strategy remains necessary in ForAda++. Specifically, we consider two variants based on the binary mask $B$: \emph{w/ $B$ (face region)}, where patches containing facial regions are assigned 1 and all other patches are assigned 0, and \emph{w/ $B$ (forged region)}, where patches containing forged regions are assigned 1 and all other patches are assigned 0. As shown in Table~\ref{tab:bb}, both variants of the masking strategy lead to performance degradation. We attribute this to the denser supervision in ForAda++, which already provides sufficient spatial focus, making the additional masking strategy unnecessary. Accordingly, we omit the masking strategy in ForAda++.
}

\begin{table}[!ht]
\caption{{Ablation study of the masking strategy in ForAda++.}}
\vspace{-0.3cm}
\setlength{\tabcolsep}{4pt} 
\small
\centering
\resizebox{\linewidth}{!}{
\begin{tabular}{c|ccc|ccc}
\toprule
\multirow{2}{*}{Setting}  & \multicolumn{3}{c|}{CDF-v2} & \multicolumn{3}{c}{DFDC}  \\ 
\cmidrule(lr){2-4} \cmidrule(lr){5-7}
&  AUC & AP & EER & AUC & AP & EER  \\
\midrule
w/ $B$ (face region) & 0.892 & 0.936 & 18.7 & 0.843 & 0.871 &23.7\\ 
w/ $B$ (forged region)& 0.890 & 0.935 & 18.9 & 0.837 & 0.867 &22.6\\
w/o B &\textbf{0.903} & \textbf{0.943} & \textbf{17.8} & \textbf{0.863} & \textbf{0.884} & \textbf{21.9}  \\
\bottomrule
\end{tabular}
}
\vspace{-0.3cm}
\label{tab:bb}
\end{table}

\smallskip
\noindent\textbf{Threshold $\nu$ on whether Facial Region Is Manipulated.}
We study the impact of the threshold $\nu$ used to determine whether a facial region $\mathcal{R}_i$ is manipulated. Specifically, we {evaluate} threshold values of 0.10, 0.15, 0.20, and 0.40. A higher threshold indicates that a region is viewed as fake only when it exhibits more discrepancies from the original, while a lower threshold makes a region more likely to be regarded as fake. Therefore, selecting an appropriate threshold requires balancing sensitivity and specificity. As reported in Table~\ref{tab:threshold}, the best performance is achieved with a moderate threshold of $\nu = 0.15$, aligning with {our} initial analysis.

\begin{table}[!ht]
\caption{Ablation Study on Threshold $\nu$.}
\vspace{-0.3cm}
\setlength{\tabcolsep}{4pt} 
\small
\centering
\begin{tabular}{c|ccc|ccc}
\toprule
\multirow{2}{*}{Threshold}  & \multicolumn{3}{c|}{CDF-v2} & \multicolumn{3}{c}{DFDC}  \\ 
\cmidrule(lr){2-4} \cmidrule(lr){5-7}
&  AUC & AP & EER & AUC & AP & EER  \\
\midrule
0.10&  0.885 & 0.934 & 19.3 &0.848 & 0.876 & 23.4  \\ 
0.15&  {0.903} & \textbf{0.943} & {17.8} &\textbf{0.863} & \textbf{0.884} & \textbf{21.9} \\ 
0.20 &   \textbf{0.906} & \textbf{0.943} & \textbf{16.9} &   0.850 & 0.871 & 22.1 \\
0.40 &  0.881 & 0.934 & 20.2 &   0.825 &0.855 & 24.7  \\
\bottomrule
\end{tabular}
\label{tab:threshold}
\end{table}

\smallskip
\noindent\textbf{Length of Learnable Prompt Suffix.}
This section investigates the effect of varying numbers of learnable context tokens. The results are shown in Table~\ref{tab:number}, where the first row represents the baseline without {the} learnable prompt suffix. A significant performance drop is observed when no learnable prompts are used, highlighting their role in effectively integrating textual information. As the number of tokens increases (\eg, to 4 and 8), performance improves gradually. {Nevertheless, adding more tokens does not always lead to performance gains}. With 16 tokens, the performance sharply drops to the level without any learnable prompts. Based on these observations, we adopt 8 tokens in the learnable prompt suffix for the main experiments.

\begin{table}[!ht]
\caption{Ablation Study on the length of learnable prompt suffix.}
\vspace{-0.3cm}
\setlength{\tabcolsep}{4pt} 
\small
\centering
\begin{tabular}{c|ccc|ccc}
\toprule
\multirow{2}{*}{Number}  & \multicolumn{3}{c|}{CDF-v2} & \multicolumn{3}{c}{DFDC}  \\ 
\cmidrule(lr){2-4} \cmidrule(lr){5-7}
&  AUC & AP & EER & AUC & AP & EER  \\
\midrule
0 & 0.879 & 0.936 & 20.4 & 0.852 & \textbf{0.886} & 22.7  \\ 
4 & 0.900& 0.941 & 18.9 &   0.860 & {0.885} &  22.0 \\
8 & \textbf{0.903} & \textbf{0.943} & \textbf{17.8} & \textbf{0.863} & {0.884} & \textbf{21.9}  \\
16 & 0.875 & 0.928 & 20.7 &   {0.853} & {0.873} & {22.9}  \\
\bottomrule
\end{tabular}
\vspace{-0.3cm}
\label{tab:number}
\end{table}

\smallskip
\noindent\textbf{Varying Loss Weight $\lambda_{0 \sim 4}$.}
We investigate the impact of varying loss weight configurations. As shown in Table~\ref{tab:loss_weight}, the effects of scaling $\lambda_0$, $\lambda_1$, $\lambda_{2}$, $\lambda_3$, and $\lambda_4$ by a factor of 10 are reported. {The results highlight that the model is sensitive to changes in the weights of $\mathcal{L}_0,\mathcal{L}_1,\mathcal{L}_4$, where improper weighting leads to a notable shift in performance. {This is because $\mathcal{L}_0$ and $\mathcal{L}_1$ are the main objective {losses} of this task, while $\mathcal{L}_4$ strongly affects the efficacy of {the} visual modality. In contrast, $\mathcal{L}_2$ and $\mathcal{L}_3$ {serve} as {auxiliary terms} and {show} limited sensitivity to weight variations.}}

\begin{table}[!ht]
\caption{Ablation study on the effect of varying loss weight $\lambda_{0 \sim 4}$.}
\vspace{-0.3cm}
\setlength{\tabcolsep}{4pt} 
\small
\centering
\begin{tabular}{c|ccc|ccc}
\toprule
\multirow{2}{*}{Loss Weights}  & \multicolumn{3}{c|}{CDF-v2} & \multicolumn{3}{c}{DFDC}  \\ 
\cmidrule(lr){2-4} \cmidrule(lr){5-7}
&  AUC & AP & EER & AUC & AP & EER  \\
\midrule
10 * $\lambda_0$& 0.855 & 0.918 & 22.7 & 0.835 &0.864 & 24.1\\ 
10 * $\lambda_1$ &0.865 & 0.924 & 21.5& \textbf{0.878} & \textbf{0.895} & \textbf{20.9} \\ 
10 * $\lambda_2$&0.895 & 0.938 &18.7& 0.848 & 0.875 &23.4 \\ 
10 * $\lambda_{3}$&0.901 & \textbf{0.945} &17.9 & 0.822& 0.875 & 21.4 \\ 
10 * $\lambda_{4}$& 0.871 &0.928 &21.0 & 0.841& 0.873 & 23.9 \\ 

- & \textbf{0.903} & 0.943 & \textbf{17.8} & 0.863 & 0.884 &21.9  \\
\bottomrule
\end{tabular}
\label{tab:loss_weight}
\end{table}

\subsection{Further Analysis}
\label{sec:Further-Analysis}
\smallskip
\noindent\textbf{Using Various VLMs. }
This part investigates the effect of using various VLMs. Specifically, we validate two VLMs: BLIP ViT-L/14~\cite{blip} {and} CLIP ViT-B/16~\cite{clip}. 
The results are shown in Table~\ref{tab:dif-Backbone}. We observe that BLIP is not suitable for the task, which {significantly} degrades the performance compared to CLIP variants. This is possible because BLIP is trained using only 129 million image-text pairs and is designed for image-text retrieval and VQA tasks with a bootstrapping mechanism. In contrast, CLIP is trained on a much larger dataset containing 400 million samples, enabling stronger zero-shot capabilities for image-based tasks. {Naturally, a larger ViT achieves better performance than the base version,} which aligns with our understanding.

\begin{table}[!ht]
\caption{Effect of using various VLMs.} 
\small%

\centering
\vspace{-0.3cm}

\begin{tabular}{c|c|cc} 
\toprule 
VLMs   & Method & CDF-v2  &  DFDC   \\
\midrule 
 BLIP ViT-L/14 \cite{blip} & ForAda & 0.607 & 0.539\\
 CLIP ViT-B/16 \cite{clip} & ForAda  & 0.837 & 0.775 \\
 CLIP ViT-B/16 \cite{clip} & ForAda++ & \textbf{0.864}& \textbf{0.786} \\
\bottomrule 
\end{tabular}
\label{tab:dif-Backbone} 
\end{table}

\smallskip
\noindent\textbf{Using Various Adapter Architectures.}
In this section, we explore the effect of different adapter architectures, including ViT-tiny (ours), ViT-small, ViT-base, and ViT-large. All adapters are trained using the same configuration as in our main experiment. The results are presented in Table~\ref{tab:adapter_s}. As shown, performance does not improve as the number of trainable parameters increases under the same training configuration. 
{This is likely due to two reasons: 1) {larger} architecture{s introduce exponentially increasing} feature dimensions. To align with CLIP, we have to {substantially} normalize the output of the adapter, likely leading to certain information loss. 2) With limited training resources, the smaller ViT-tiny architecture can be fully trained to capture forgery traces, whereas the larger architectures do not show additional benefits.}

\begin{table}[!ht]
\caption{Effect of using various adapter structures.}
\centering
\small%
\vspace{-0.3cm}
\begin{tabular}{c|c|ccc} 
\toprule 
Adapter & Method & CDF-v2 &  DFDC  & DFD \\
\midrule 
Large & \multirow{4}{*}{ForAda} & 0.891 & 0.832  & 0.905  \\
Base  & & 0.869 & 0.833  & 0.902\\
Small & & 0.886 & 0.825  & 0.906 \\
Tiny & & \textbf{0.900} & \textbf{0.843}  & \textbf{0.933} \\
\midrule
Large & \multirow{4}{*}{ForAda++} & 0.872  & \textbf{0.866} & 0.937  \\
Base  & & 0.867 &  0.845& 0.933\\
Small & & 0.880 & 0.860&  \textbf{0.940}  \\
Tiny  & &  \textbf{0.903} &  {0.863} &  {0.922}\\
\bottomrule 
\end{tabular}
\label{tab:adapter_s} 
\end{table}

\smallskip
\noindent\textbf{Effect of Integrating Visual Features into Learnable Prompt Suffix.}
This section evaluates the impact of injecting visual features into the learnable prompt suffix. Specifically, we perform attention-based fusion between the visual features and learnable prompt tokens, which is then added to the original learnable prompt tokens and passed through a feed-forward network (FFN) to generate the final text prompt.
We explore three types of visual features for fusion: the refined CLS* tokens ($\bm{X}_{\mathrm{[CLS^*]}}$), the standard CLS token ($\bm{X}_{\mathrm{[CLS]}}$), and visual tokens ($\bm{X}_{\mathrm{[vis]}}$). As shown in Table~\ref{tab:vtc}, the best performance is achieved when visual feature fusion is not applied. This is possible because fusing with visual features directly can introduce redundancy, hindering prompt tuning.

\begin{table}[!ht]
\caption{Ablation study on the effect of integrating visual features into learnable prompt suffix.}
\vspace{-0.3cm}
\setlength{\tabcolsep}{4pt} 
\small
\centering
\begin{tabular}{c|ccc|ccc}
\toprule
\multirow{2}{*}{Feature Type}  & \multicolumn{3}{c|}{CDF-v2} & \multicolumn{3}{c}{DFDC}  \\ 
\cmidrule(lr){2-4} \cmidrule(lr){5-7}
&  AUC & AP & EER & AUC & AP & EER  \\
\midrule
$\bm{X}_{\mathrm{[CLS^*]}}$& 0.874&0.925 &  21.4 & 0.829 & 0.848 & 24.8\\ 
$\bm{X}_{\mathrm{[CLS]}}$ &0.874 &0.914 & 20.9 &  0.816 & 0.823 & 26.0\\
$\bm{X}_{\mathrm{[vis]}}$ &0.887& 0.932 & 19.4 & 0.835 & 0.853 &24.6 \\
w/o fusion &  \textbf{0.903} &  \textbf{0.943} & \textbf{17.8} &  \textbf{0.863} &  \textbf{0.884} &  \textbf{21.9}   \\

\bottomrule
\end{tabular}
\label{tab:vtc}
\end{table}

\smallskip
\noindent\textbf{Other Fine-tuning Strategies with Textual Modality.}
We further investigate alternative fine-tuning strategies for CLIP, including: (1) freezing the entire CLIP model, (2) fine-tuning the top three layers of CLIP’s visual encoder using LoRA, (3) fine-tuning only the ln1 and ln2 normalization layers across all visual encoder layers via the subset method, (4) appending a lightweight adapter module after the CLIP visual encoder, and (5) integrating our ForAda structure with the text encoder of CLIP while removing the classification loss $\mathcal{L}_0$ based on $\bm{X}_{\mathrm{[CLS^*]}}$. For all approaches, the text prompt is standardized as “This is a real/fake person”, and contrastive learning is applied between the visual and textual embeddings to determine image authenticity.
Results are reported in Table~\ref{tab:other_fine}. It is observed that our methods consistently outperform the alternatives, highlighting the effectiveness of
task-specific adaptation tailored to face forgery detection.

\begin{table}[!ht]
\caption{Comparison of different fine-tuning strategies, with evaluation metrics being frame-level AUC.}  
\small%
\centering
\vspace{-0.3cm}

\begin{tabular}{c|ccc} 
\toprule 
Method   & CDF-v2  &  DFDC &  DFD  \\
\midrule 
Frozen & 0.378 &  0.529 &  0.514 \\
LoRA &0.795  &0.749 & 0.842 \\
Subset & 0.767 & 0.788 & 0.887\\
Naive Adapter & 0.724  & 0.720 & 0.803\\
ForAda (w/o $\mathcal{L}_0$)& 0.817  & 0.798 & 0.853\\
\midrule 
\rowcolor{hl}{ForAda++}  & \textbf{0.903} & \textbf{0.863} &  \textbf{0.922} \\
\bottomrule 
\end{tabular}
\label{tab:other_fine} 
\end{table}

\smallskip
\noindent{\textbf{Analysis of Bidirectional Interaction Mechanism.}}
{
This section investigates the effect of the bidirectional interaction mechanism in ForAda and ForAda++, with results summarized in Table~\ref{tab:biim}. Removing Adapter$\rightarrow$CLIP, the key pathway that guides CLIP attention toward forgery-relevant regions, disrupts attention modulation. Therefore, in the \emph{w/o Adapter$\rightarrow$CLIP} setting, we replace the adapter-derived attention bias with a learnable vector of the same dimension, resulting in a substantial performance drop. In the \emph{w/o CLIP$\rightarrow$Adapter} setting, the fusion of CLIP features into the shallow layers of the adapter is removed, leading to a milder yet consistently negative performance trend. Removing both directions further degrades performance. These results indicate that Adapter$\rightarrow$CLIP feeds task-specific forgery knowledge into CLIP, while CLIP$\rightarrow$Adapter provides complementary visual priors, and the two directions together facilitate effective knowledge exchange. In addition, ForAda shows a more pronounced performance drop than ForAda++ under these settings, which we attribute to the textual modality introduced in ForAda++, as it provides additional semantic information that helps stabilize performance and improve robustness.
}

\begin{figure}[!ht]
    \centering
    \includegraphics[width=0.82\linewidth]{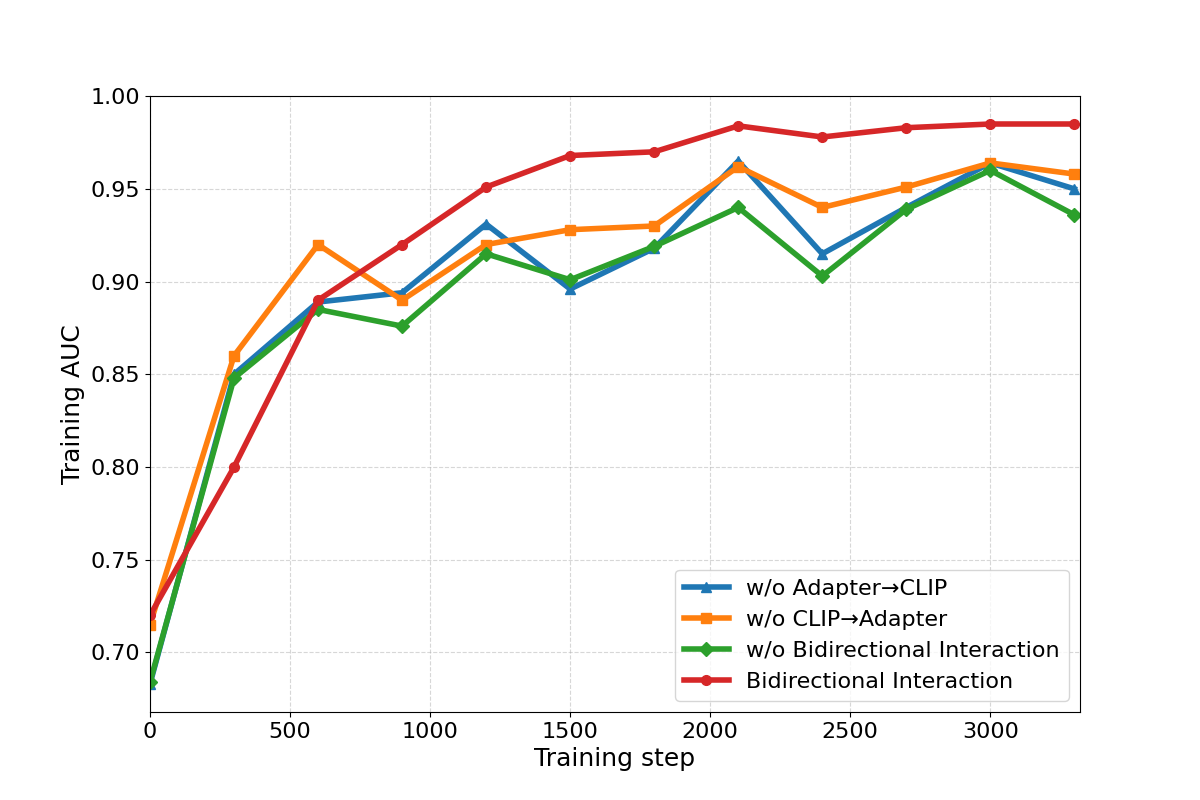}
    \caption{{
    {Training AUC evolution across different interaction configurations. The proposed bidirectional interaction leads to more stable optimization and higher learning efficiency.}}
    }
    \label{fig:learn_eff}
\end{figure}

{We further compare the evolution of training AUC for different ForAda variants to evaluate learning efficiency. As shown in Fig.~\ref{fig:learn_eff}, the full bidirectional interaction model exhibits more stable optimization and a higher performance ceiling, whereas removing either interaction pathway leads to larger oscillations and inferior performance. This further confirms that the two directions are complementary and jointly contribute to improvements in performance, training stability, and learning efficiency.}

\begin{table*}[!ht]
\caption{{{Effect of the Bidirectional Interaction Mechanism.}}}
\label{tab:biim}
\vspace{-0.3cm}
\centering
\small
\begin{tabular}{c|c|ccc} 
\toprule 
Adapter & Method & CDF-v2 & DFDC & DFD \\
\midrule 

{w/o Adapter}$\rightarrow$CLIP & \multirow{4}{*}{ForAda}& 0.851 & 0.792 & 0.913 \\
w/o CLIP$\rightarrow$Adapter &  & 0.885 & 0.841 & 0.914 \\

w/o Bidirectional Interaction & & 0.859 & 0.783 & 0.899 \\
Bidirectional Interaction& & \textbf{0.900} & \textbf{0.843} & \textbf{0.933} \\
\midrule

w/o Adapter$\rightarrow$CLIP & \multirow{4}{*}{ForAda++}& 0.889 & 0.804 & 0.916 \\
w/o CLIP$\rightarrow$Adapter &  & 0.900 & 0.853 &  \textbf{0.922} \\
w/o Bidirectional Interaction & & 0.883 & 0.801 & 0.915 \\
Bidirectional Interaction & & \textbf{0.903} & \textbf{0.863} & \textbf{0.922} \\
\bottomrule 
\end{tabular}
\end{table*}

\begin{figure*}[!ht]
    \centering
    \includegraphics[width=1\linewidth]{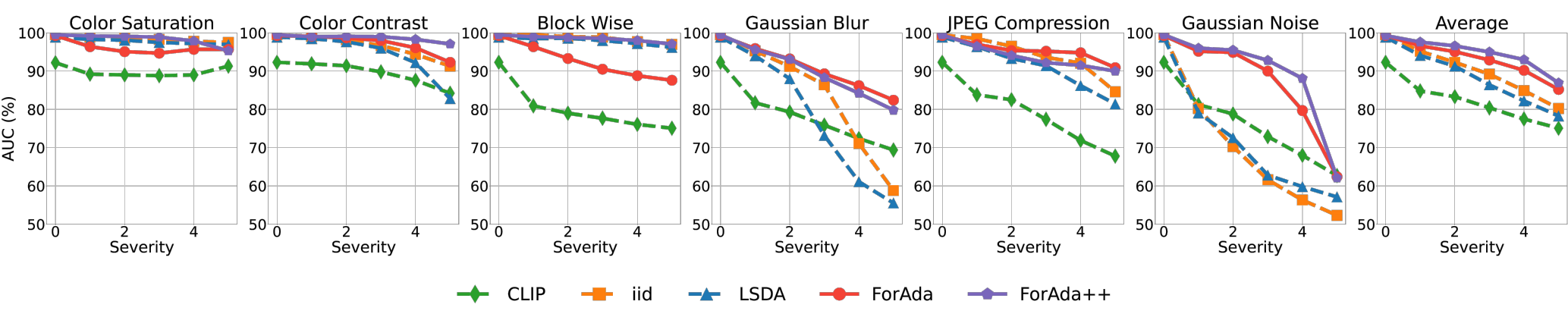}
    \vspace{-0.8cm}
    \caption{Robustness Analysis. Our method is compared with CLIP~\cite{clip}, IID~\cite{iid}, and LSDA~\cite{lsda} across five levels of six particular types of perturbations in video-level AUC.}
    \label{fig:robustness}
    \vspace{-0.3cm}
\end{figure*}

\begin{table*}[!ht]
\caption{Performance of detecting whole face images synthesized by GANs and Diffusion models. }
\setlength{\tabcolsep}{4pt} 
\small%
\centering
\vspace{-0.3cm}
\begin{tabular}{c|ccc|ccc|ccc|ccc|ccc} 
\toprule 
\multirow{2}{*}{Method}  & \multicolumn{3}{c|}{StarGAN \cite{stargan}} & \multicolumn{3}{c|}{CramerGAN \cite{gan_cramer}} & \multicolumn{3}{c|}{MMDGAN \cite{gan_mmd}} & \multicolumn{3}{c|}{ADM \cite{difussion_adm}}  & \multicolumn{3}{c}{PNDM \cite{difussion_pndm}}
 \\ 
\cmidrule(lr){2-4} \cmidrule(lr){5-7} \cmidrule(lr){8-10} \cmidrule(lr){11-13} \cmidrule(lr){14-16}  
 &  AUC & AP & EER &  AUC & AP & EER  &  AUC & AP & EER  &  AUC & AP & EER &  AUC & AP & EER   \\
 \midrule
 IID~\cite{iid} & 0.699 & 0.697 & 34.7& 0.639 & 0.617 &40.3 & 0.565 & 0.552 & 45.3 & 0.609 & 0.617& 42.0& 0.279& 0.391& 63.3\\
LSDA~\cite{lsda} & 0.772 & 0.796  &  29.0 &  0.675  &0.646& 35.8 & 0.621 & 0.592 & 40.3 &0.577  & 0.549 & 42.8 &0.572&0.546&40.6 \\
\midrule
\rowcolor{hl} ForAda & 0.954 & 0.950  & \textbf{10.9} &  0.936  & 0.945 & 13.7 &  0.896 &{0.911}& {18.6} &  \textbf{0.806} &\textbf{0.793} & \textbf{26.3} &0.638 & {0.624} &{39.3}\\
\rowcolor{hl} ForAda++  & \textbf{0.959} & \textbf{0.960} &\textbf{10.9} & \textbf{0.972} &\textbf{0.972}&\textbf{8.75} &\textbf{0.960}& \textbf{0.961}&\textbf{10.6} &0.781 &0.769 &28.7 &\textbf{0.667} &\textbf{0.656}&\textbf{37.8} \\ 
\bottomrule 
\end{tabular}
\label{tab:whole} 
\end{table*}

\smallskip
\noindent\textbf{Robustness.}
This part studies the performance of our method {under} various perturbations. We employ the same perturbation {configuration} as in~\cite{DeeperForensics}, which includes five levels of six different types of perturbations: color saturation, color contrast, block wise, gaussian blur, jpeg compression, {and} gaussian noise. We compare our two methods with LSDA, IID, and vanilla CLIP methods {using} video-level AUC. The results are shown in Fig.~\ref{fig:robustness}. It can be observed that our methods demonstrate greater robustness compared to the others. Notably, after integrating the text modality, the model enhances its resistance to perturbations in the visual modality. All methods are trained on the FF++ c23 dataset.

\smallskip
\noindent\textbf{Detecting Whole Face Image Synthesis.}
Although our methods are restricted to detecting face forgeries, we further explore their potential for detecting whole face images synthesized by GANs and Diffusion models. Specifically, we select three GANs, including StarGAN~\cite{stargan}, CramerGAN~\cite{gan_cramer}, and MMDGAN~\cite{gan_mmd}, and two Diffusion models, including ADM~\cite{difussion_adm} and PNDM~\cite{difussion_pndm}. Table~\ref{tab:whole} shows the performance of our method compared to IID and LSDA. The results indicate that, despite being designed specifically for face forgery tasks, our method exhibits satisfactory performance and still surpasses the counterparts by a large margin on both GANs and Diffusion models.
Moreover, ForAda++ further improves performance over ForAda, highlighting the efficacy of the textual modality even under broader forgery distributions.

\begin{figure*}[!ht]
    \centering
    \includegraphics[width=1\linewidth]{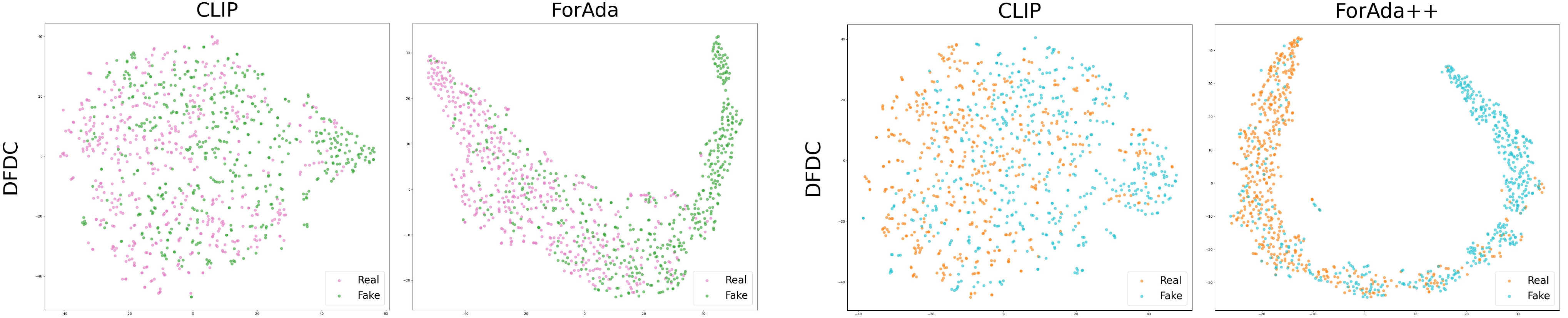}
    \vspace{-0.8cm}
    \caption{
    T-SNE Visualizations. The left figure compares the feature distributions of the ForAda method and CLIP, while the right figure presents the corresponding comparison between the ForAda++ method and CLIP.
    }
    \label{fig:tsne}
\end{figure*}

\smallskip
\noindent\textbf{Complexity Analysis.}
Since we freeze the CLIP model and only train the lightweight adapter, we significantly reduce the training overhead to 5.7M parameters.
 Although we introduce a textual branch in ForAda++, we add only a single-digit number of context tokens, resulting in an almost negligible increase in parameter count.
Table~\ref{tab:complexity} compares the parameter counts and video-level AUC performance of various methods, all trained on the FF++ c23 dataset. Notably, other methods require three to six times more trainable parameters than ours, yet our method achieves a substantial performance improvement, highlighting the effectiveness of our method.

\begin{table}[!ht]
\caption{Complexity analysis in video-level AUC.} 
\small
\centering
\vspace{-0.3cm}
\begin{tabular}{c|c|c|cc} 
\toprule 
Method  & \# Params  & CDF-v2 & DFDC  \\
\midrule 
LipForensics \cite{lips} & 36.0M & 0.824 & 0.735 \\
FTCN \cite{ftcn} & 26.6M & 0.869 & 0.740 \\
DCL \cite{dcl} & 19.35M & 0.823 & 0.767 \\
CFM \cite{cfm} & 25.37M & 0.897 & 0.802 \\
\midrule
\rowcolor{hl} ForAda & \textbf{5.7M} & \textbf{0.957} & {0.872}  \\
\rowcolor{hl} ForAda++ & \textbf{5.7M} & {0.940} & \textbf{0.887}  \\

\bottomrule 
\end{tabular}
\label{tab:complexity} 
\end{table}


\smallskip
\noindent\textbf{T-SNE Visualization.}
To better demonstrate the efficacy of our methods, we visualize the output of CLIP before and after using the adapter on the DFDC dataset. Specifically, we randomly select 500 real and 500 fake faces from the testing set and employ the T-SNE algorithm~\cite{tsne} for visualization. The visualization results are shown in Fig.~\ref{fig:tsne}, with ForAda on the left and ForAda++ on the right. It can be observed that the original CLIP embeddings fail to provide a clear separation between real and fake samples in this challenging setting. In contrast, ForAda achieves substantially better feature separation. Moreover, the incorporation of the textual modality in ForAda++ further amplifies this effect, yielding more distinct clustering between real and fake faces. These results underscore the capacity of our methods to effectively capture generalizable forgery traces.

\begin{figure*}[!ht]
    \centering
    \includegraphics[width=0.65\linewidth]{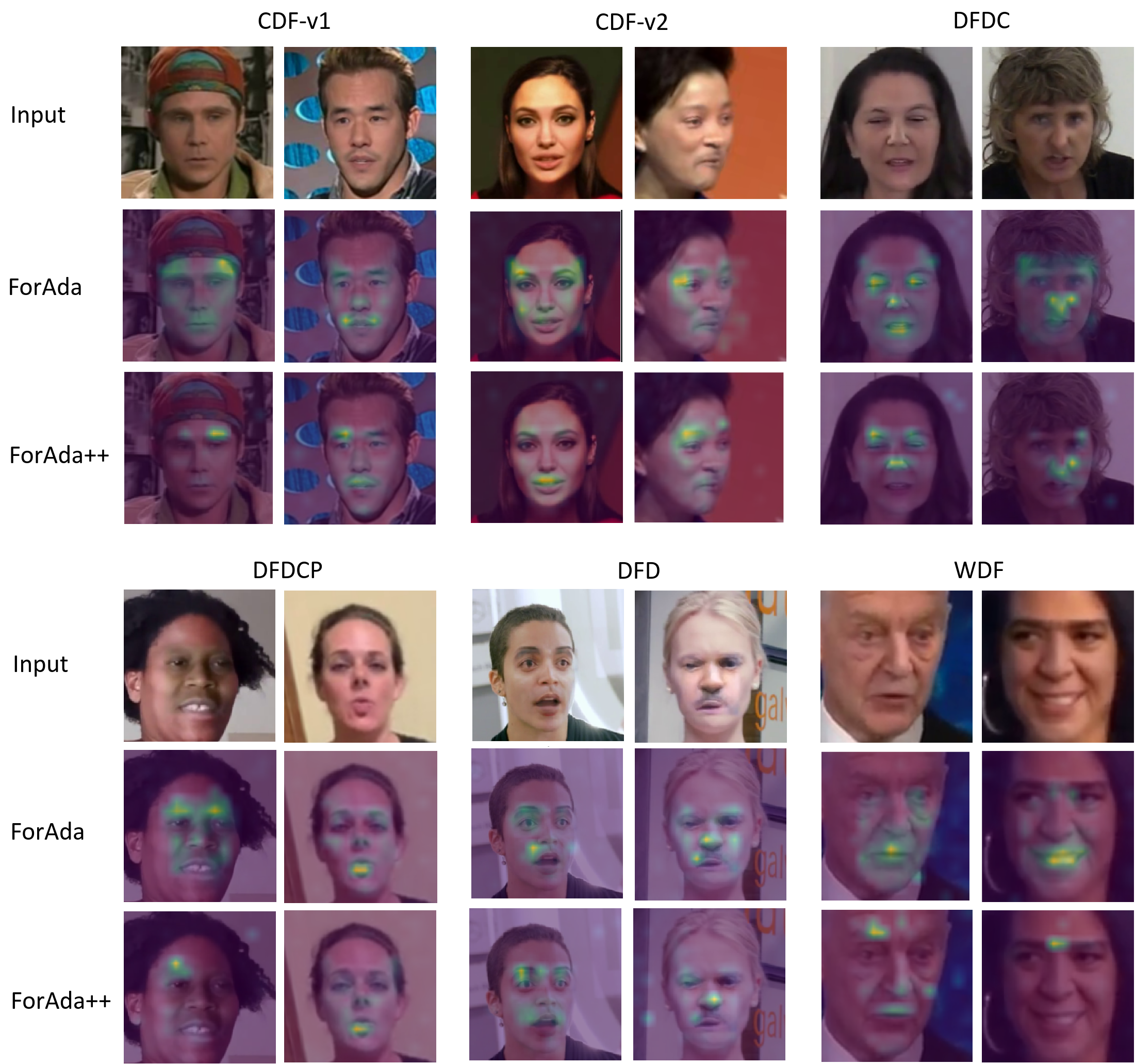}
    \caption{{
    Attention visualizations of {ForAda} and {ForAda++} on six datasets. Heatmaps are generated from gradient-weighted attention responses from CLS* tokens to visual tokens with respect to the fake prediction.
    }}
    \label{fig:grad-cam}
\end{figure*}

\smallskip
\noindent{\textbf{Attention Visualization.}}
{
To evaluate whether ForAda and ForAda++ can consistently capture fake facial cues across different forgery types, we visualize the class-specific responses in the last attention layer of CLIP’s visual encoder on multiple test datasets. Specifically, the heatmaps are generated from gradient-weighted attention responses from CLS* tokens to visual tokens with respect to the fake prediction. As shown in Fig.~\ref{fig:grad-cam}, the heatmaps exhibit broadly consistent patterns across different forgery types. ForAda tends to focus on the entire forged face region, highlighting cues such as blending boundaries and flickering artifacts around the eyes, nose, and mouth, which demonstrates its consistent ability to capture forgery-related cues across different manipulations. With forgery-aware prompt learning, ForAda++ further focuses on the most discriminative local regions, such as the eyebrow regions in CDF-v1, CDF-v2, and WDF.
}

\smallskip
\noindent\textbf{Limitations.}
Our method is designed specifically for face forgery detection, rather than whole face image synthesis detection. Thus, it only shows decent performance on recent GAN or Diffusion model-generated faces. In future work, we plan to further explore the capabilities of CLIP and aim to develop a more universal detector that can effectively handle both face forgery and whole face synthesis detection.

\section{Conclusion}
In this paper, we introduce \textbf{Forensics Adapter}, an adapter network designed to transform CLIP into a generalizable face forgery detector. 
Different from existing methods, our adapter is specifically tailored to capture the unique blending boundaries characteristic of forged faces using task-specific objectives. We then introduce an interactive strategy that enables CLIP and the adapter to collaboratively focus on face forgery traces. Notably, our approach to adapting CLIP for face forgery detection is novel, addressing a gap that previous methods have not explored. 
{Extensive experiments demonstrate that, with only $\bm{5.7M}$ trainable parameters, our adapter achieves substantial performance gains across six standard datasets.}
Building on this foundation, we further present \textbf{Forensics Adapter++}, which incorporates the textual modality via a newly proposed forgery-aware prompt learning strategy, resulting in an additional performance boost of $\bm{1.3\%}$. We believe these contributions provide a strong foundation for future CLIP-based face forgery detection research.


{
   \bibliographystyle{IEEEtran}
   \bibliography{references}
}
\end{document}